\documentclass[article]{agujournal2019}
\usepackage{url} %this package should fix any errors with URLs in refs.
\usepackage{lineno}
\usepackage{soul}
\usepackage{dsfont}
\usepackage{gensymb}
\usepackage{subcaption}
\usepackage{hyperref}

\RequirePackage{apacite}
\let\cite\shortcite %xx So get et al. with three authors the first time.
\let\citeA\shortciteA %xx Ditto.

\usepackage{cleveref}

\draftfalse

\journalname{a journal (non-peer reviewed draft)}

\begin{document}

%% ------------------------------------------------------------------------ %%
%  Title
%
% (A title should be specific, informative, and brief. Use
% abbreviations only if they are defined in the abstract. Titles that
% start with general keywords then specific terms are optimized in
% searches)
%
%% ------------------------------------------------------------------------ %%

% Example: \title{This is a test title}

\title{Tackling water table depth modeling via machine learning: From proxy observations to verifiability}

%% ------------------------------------------------------------------------ %%
%
%  AUTHORS AND AFFILIATIONS
%
%% ------------------------------------------------------------------------ %%

% Authors are individuals who have significantly contributed to the
% research and preparation of the article. Group authors are allowed, if
% each author in the group is separately identified in an appendix.)

% List authors by first name or initial followed by last name and
% separated by commas. Use \affil{} to number affiliations, and
% \thanks{} for author notes.
% Additional author notes should be indicated with \thanks{} (for
% example, for current addresses).

% Example: \authors{A. B. Author\affil{1}\thanks{Current address, Antartica}, B. C. Author\affil{2,3}, and D. E.
% Author\affil{3,4}\thanks{Also funded by Monsanto.}}

\authors{Joseph Janssen\affil{1}, Ardalan Tootchi\affil{1}, and Ali A. Ameli\affil{1}}

% \affiliation{1}{First Affiliation}
% \affiliation{2}{Second Affiliation}
% \affiliation{3}{Third Affiliation}
% \affiliation{4}{Fourth Affiliation}

\affiliation{1}{Department of Earth, Ocean, and Atmospheric Sciences, University of British Columbia, Vancouver, BC, Canada}

%(repeat as many times as is necessary)

%% Corresponding Author:
% Corresponding author mailing address and e-mail address:

% (include name and email addresses of the corresponding author.  More
% than one corresponding author is allowed in this LaTeX file and for
% publication; but only one corresponding author is allowed in our
% editorial system.)

% Example: \correspondingauthor{First and Last Name}{email@address.edu}

\correspondingauthor{Joseph Janssen}{joejanssen@eoas.ubc.ca}

%% Keypoints, final entry on title page.

%  List up to three key points (at least one is required)
%  Key Points summarize the main points and conclusions of the article
%  Each must be 140 characters or fewer with no special characters or punctuation and must be complete sentences

%% ------------------------------------------------------------------------ %%
%
%  ABSTRACT and PLAIN LANGUAGE SUMMARY
%
% A good Abstract will begin with a short description of the problem
% being addressed, briefly describe the new data or analyses, then
% briefly states the main conclusion(s) and how they are supported and
% uncertainties.

% The Plain Language Summary should be written for a broad audience,
% including journalists and the science-interested public, that will not have 
% a background in your field.
%
% A Plain Language Summary is required in GRL, JGR: Planets, JGR: Biogeosciences,
% JGR: Oceans, G-Cubed, Reviews of Geophysics, and JAMES.
% see http://sharingscience.agu.org/creating-plain-language-summary/)
%
%% ------------------------------------------------------------------------ %%

%% \begin{abstract} starts the second page

\begin{abstract}
Spatial patterns of water table depth (WTD) play a crucial role in shaping ecological resilience, hydrological connectivity, and human-centric systems. Generally, a large-scale (e.g., continental or global) continuous map of static WTD can be simulated using either physically-based (PB) or machine learning-based (ML) models. We construct three fine-resolution (500 m) ML simulations of WTD, using the XGBoost algorithm and more than 20 million real and proxy observations of WTD, across the United States and Canada. The three ML models were constrained using known physical relations between WTD's drivers and WTD and were trained by sequentially adding real and proxy observations of WTD. Through an extensive (pixel-by-pixel) evaluation across the study region and within ten major ecoregions of North America, we demonstrate that our models (corr=0.6-0.75) can more accurately predict unseen real and proxy observations of WTD compared to two available PB simulations of WTD (corr=0.21-0.40). However, we still argue that currently-available large-scale simulations of static WTD could be uncertain within data-scarce regions such as steep mountainous regions. We reason that biased observational data mainly collected from low-elevation floodplains and the over-flexibility of available models can negatively affect the verifiability of large-scale simulations of WTD. Ultimately, we thoroughly discuss future directions that may help hydrogeologists decide how to improve machine learning-based WTD estimations. In particular, we advocate for the use of proxy satellite data, the incorporation of physical laws, the implementation of better model verification standards, the development of novel globally-available emergent indices, and the collection of more reliable observations.
\end{abstract}

\noindent%
{\it Keywords:}  
%3 to 6 keywords, that do not appear in the title
Machine learning, physically-based models, Groundwater, Water Table Depth, North America, Ecoregions, Model uncertainty, Observation uncertainty

\section{Introduction} 

Groundwater is the most abundant source of liquid freshwater on Earth, with an important influence on above-ground processes \cite{gleeson2016global}. Groundwater affects water quantity and quality in streams \cite{miguez2012role,ali2011linking}, lakes \cite{vaheddoost2018interaction,xu2021investigating,ameli2014semianalytical}, and wetlands \cite{ameli2019groundwaters,ameli2017quantifying}, and regulates the behaviour of ecological and human-centric systems \cite{siebert2010groundwater,huggins2023groundwater}. In riverine ecosystems, groundwater provides critical water supplies to plants in arid riparian and non-riparian areas, without which trees and plants would quickly deteriorate in drought conditions \cite{kibler2021brown,xu2022impact}. The central property of groundwater at a particular location and the particular aquifer is the water table depth (WTD), defined as the depth below which the ground becomes fully saturated with water. Static WTD, which is the main focus of this paper, is defined as the long-term average (non-transient) water table depth at a given location. This quantity can vary by several orders of magnitude from one location to another. Although transient WTD simulations are often needed to form robust conclusions about hydrological processes, large-scale (e.g., continental and global scales) fine-resolution spatial estimates of static WTD, can be used as a good first-order approximation in order to inform (1) the direction of groundwater movement \cite{freeze1979}, including whether or not the roots of trees in a given location have access to water \cite{tai2018distributed,cooper2003physiological,schook2022riparian}, (2) the spatial heterogeneity of hydrologic processes within and across catchments as well as the celerity of the rainfall-streamflow response \cite{scaini2018hillslope}, and the nonlinearity of stormflow \cite{kidron2021comparing} and baseflow generation \cite{ali2011linking,detty2010threshold,kim2004throughflow, li2023upland}, (3) the overall gaining or losing status of surface water bodies connected to groundwater systems \cite{jasechko2021widespread}, (4) the delineation of catchment boundaries \cite{hinton1993physical}, and (5) the spatial extent of drought and flood vulnerabilities \cite{brooks2015hydrological,smith2020groundwater,gleeson2021gmd}.

Previous studies have attempted to simulate static WTD at local, regional, national, and global scales using various methods \cite{zell2020calibrated,maxwell2015high,fan2013global,koch2019modelling}. For example, in an aquifer near Toronto, Canada, \citeA{desbarats2002kriging} used a geostatistical method (kriging with external drift) to estimate the WTD at 90m resolution using the information of elevation and topographic wetness index. \citeA{kollet2008capturing} and \citeA{tran2022hydrological} used ParFlow, a physically-based groundwater model, to simulate WTD in a catchment in Oklahoma and Colorado, respectively. Further, national-scale simulations of WTD have been made by \citeA{maxwell2015high} using ParFlow (across a large portion of the United States with 1 km resolution), by \citeA{koch2021high} using a machine learning algorithm called catboost (across Denmark with 10 m resolution), by \citeA{ben2023mapping} using kriging (across Spain and Portugal with 4 km resolution), and by \citeA{koch2019modelling} using random forests to simulate winter-time static WTD in Denmark with 50 m resolution. One of the most prominent examples of a global simulation of WTD was conducted by \citeA{fan2013global}, who used a physically-based model to develop a 1 km grid of WTD for the entire planet. Another estimation of global WTD was made by \citeA{de2017global} using PCR-GLOBWB and MODFLOW at a 10 km resolution, as an extension of their earlier work \cite{de2015high}. As an improvement to \citeA{de2017global}, \citeA{verkaik2024globgm} developed the PCR-GLOBWB-MODFLOW global groundwater model to simulate WTD at 1 km resolution. Other global physically-based models of WTD include simulations from \citeA{zeng2018global} (1 km resolution), \citeA{reinecke2019challenges} (9 km resolution), \citeA{niu2007development} (100 km resolution), and \citeA{lo2011precipitation} (300 km resolution).

% For example, in an aquifer near Toronto, Canada, \citeA{desbarats2002kriging} used a geostatistical method (kriging with external drift) to estimate the WTD at 90m resolution using elevation and topographic wetness index data. They tested both of these variables because previous works have observed that the water table is a damped mirror image of local topography, and physically-based rainfall-runoff models such as TOPMODEL relate WTD and topographic index linearly. Elevation was found to be a better predictor of WTD compared to topographic index, though it was argued that the latter is more physically meaningful. Similarly, \citeA{ahmadi2008application} used kriging and cokriging to estimate WTD in a basin in Iran during wet, average, and dry conditions. 

% "For North America, Fan et al. (2007) and Miguez-Macho et al. (2007) linked a land surface model with a two-dimensional gradient-based GW model and computed, with a daily time step, GW flow, water table elevation, GW–SW interaction, and capillary rise, using a spatial resolution of 1.25km. One challenge was the determination of the river conductance that affects the degree of GW–SW interaction. A computationally very expensive integrated simulation of dynamic SW, soil, and GW flow using Richards' equation for variably saturated flow was achieved at a spatial resolution of 1km for the continental US by applying the ParFlow model (Maxwell et al., 2015). In both studies, GW abstractions were not taken into account."-\cite{reinecke2019challenges}

Attempts to simulate national or global-scale WTD have typically fallen into two categories: (1) physically-based models and (2) machine learning models, with each containing notable shortcomings. Physically-based numerical methods are flexible enough to incorporate spatiotemporal variations in recharge, heterogeneous soil properties, and time-varying boundary conditions. However, they may only make valid predictions for locations that perfectly follow a long list of assumptions under which the model was built (i.e., laboratory scale equations, laboratory conditions, homogeneity, laminar flow, etc...), with the scale being one of the most important and often contradicted assumptions \cite{herrera2023estimation,beven1996discussion}. Darcy's law and Richards' equation form the backbone of physically-based models, but they may lose their validity when applied beyond laboratory scales and conditions (e.g., the scale of 1-hectare hillslope) \cite{mcdonnell2007moving}. This implies that, due to nonlinearity and heterogeneity, a model setup with a fine grid resolution and important parameters such as effective hydraulic conductivity, which is accurately measured at the same fine resolution, is required to reliably and robustly apply physically-based models \cite{wildemeersch2014assessing,vogel2008estimation,beven1995linking,beven2001far}. See \citeA{beven1996discussion} for a critique of distributed models that violate these scale assumptions. Attempting to run physically-based models at much smaller scales would be challenging due to dramatically increased computational costs and the lack of accurate forcing/input data at such scales. Without major breakthroughs in computing, running a physically-based model at such a fine resolution for an entire continent may continue to be computationally infeasible \cite{koch2021high}. This, in addition to the unavailability of parameter datasets at such fine scales, may partially explain why global-scale physically-based simulations of WTD generally have coarse resolutions (e.g., 1 km in \citeA{fan2013global} and 10 km in \citeA{de2015high}). The vertical extent of the simulation domain in physically-based models can also affect the accuracy with which WTD can be estimated. \citeA{maxwell2015high} stated that one of the main limitations of their physically-based model developed to estimate WTD across a large portion of the USA was specifying the lower model boundary at 102 m. The lack of knowledge on the proper location of a model's lower boundary condition leads to possible underestimation of WTD magnitude in dry and mountainous regions, where it is known that actively circulating groundwater can be as deep as 200 meters \cite{condon2020bottom}. Moreover, physically-based simulations strongly depend on parameters such as hydraulic conductivity, which are challenging to estimate accurately at global scales with fine resolution \cite{de2015high,maxwell2015high,ofterdinger2014hydraulic}. While parameter uncertainty also exists in machine learning simulations — as will be discussed below — machine learning models can manage this uncertainty by down-weighting the influence of uncertain parameters.  

Machine learning methods may resolve some of the previously mentioned shortcomings of physically-based models. Machine learning models (1) do not rely on uncertain and scale-dependent physically-based assumptions, (2) are pronouncedly less computationally intensive, (3) will down-weight important variables with high measurement uncertainty such as hydraulic conductivity, and (4) have shown to provide appropriate predictive accuracy once compared against WTD observational data. The latter is particularly important, as large-scale physically-based simulations of WTD have been mostly verified against estimated hydraulic heads and not against direct observations of WTD (see the discussions in \citeA{reinecke2020importance} and \citeA{yang2023high} on how such model validation can hinder the development of trustworthy models). We especially emphasize this point because, even after the clear discussion of \citeA{reinecke2020importance}, several works have continued to validate their large-scale groundwater models against the estimated hydraulic head   \cite{de2020hyper,rahman2023simulating,callaghan2024water}. In contrast, when machine learning models were used to simulate WTD, such as in \citeA{koch2019modelling}, \citeA{koch2021high}, and \citeA{ma2024water}, the performance of the model was verified by comparing the modelled versus observed WTD observations. 

Despite the advantages of machine learning in modelling large-scale static WTD, such models strongly rely on observational data, which, in groundwater hydrology, have considerable bias, uncertainty and sparsity issues. The lack of physical realism, together with sparse, biased, and uncertain observational data, reduces the chance that machine learning models will learn the \textit{true} functional (or structural) behavior of groundwater systems. This is critically required to generalize and robustly extrapolate WTD estimates beyond available sparse observations. Poor measurement quality, sparse well drilling in certain areas, unknown impacts of external and non-natural drivers such as pumping, vertical gradients due to having an unknown mix of measurements from confined, unconfined, and perched aquifers may lead to spurious relationships in WTD datasets that the machine learning model will hopelessly follow \cite{koch2021high}. One other notable downside of current ML models in hydrology is that they only aim to predict a single variable (i.e., WTD), however, models such as ParFlow can simulate/validate both WTD and surface flow simultaneously \cite{maxwell2015high}. Still, the main objective of some of the studies that use physically-based models is the large-scale simulation of WTD \cite{fan2013global,fan2007incorporating,de2015high}, and we continue with this trend. Of course, the limitations of ML models are not limited to those noted above. More limitations and sources for uncertainty will be extensively discussed in Section \ref{sec:discussion}.

With the advent of technologies such as geophysical surveys and remote sensing products, we are now amassing fine-resolution globally available data that can (in)directly inform WTD. One notable example includes a fine-resolution (e.g., 30 m) satellite-based map of the probability of surface water occurrence \cite{pekel2016high}, obtained through merging time-varying satellite information. These types of fine-resolution proxy data have rarely been used in developing (or validating) machine learning or physically-based groundwater models. Furthermore, machine learning models are becoming more flexible in incorporating physical-based information, which can be used to increase the physical realism of such models.

In this paper, we estimate three fine-resolution (500 m) spatially continuous maps of (static) WTD in unconfined aquifers across the United States and Canada. We leverage the computational efficiency and flexibility of machine learning while incorporating physical boundary conditions, known to constrain groundwater systems, and physical (functional) relationships known to control WTD. In doing so, we use (1) more than 9 million real observations of WTD collected at more than 900,000 observational wells across the United States and Canada, (2) a physically-constrained XGBoost machine learning model, and (3) more than 12 million satellite-based proxy observations. Using the XGBoost machine learning model and subsets of real and proxy observations, we simulate three sets of simulated WTD across the study region. To obtain these three sets of predictions, the training data used to develop each machine learning model increases from model one to model three by sequentially adding satellite-based proxy observations to real observations of WTD. In doing so, we showcase the consequences of using proxy observations in simulating WTD. Ultimately, using independent/unseen real and proxy observations of WTD, we evaluate and compare our three machine-learning simulations of WTD across ten ecoregions of North America. Additionally, in this comparison, we include two available physically-based simulations of static WTD, which cover our study regions and our study period. These inter-model and model-observation comparisons indicate where different machine learning and physically-based simulations corroborate each other's predictions of WTD and where each simulation follows (or does not follow) currently available observations. Finally, focusing on structural drawbacks, prediction verifiability, and observational uncertainties, we critically discuss the challenges of machine learning-based modelling of WTD while aiming to inform future model development, data collection, and experimental analysis. In doing so, this paper goes beyond the previous machine learning-based studies of WTD \cite{ma2024water,koch2019modelling,koch2021high}, by (1) training three physically-constrained machine learning models based on multiple sources of (proxy) observations and systematically evaluating how the addition of new proxy observations affect performance and reliability (2) comparing the predictive performance of these three machine learning models and two available physically-based simulations across ten eco-regions of North America, and (3) discussing the verifiability of these models.

\section{Data}
\label{sec:data}

\subsection{Input Variables Used to Develop Machine Learning Models}
\label{sec:inputdata}
\subsubsection{Climate data}
\label{sec:inputdata_climate}
Climatic attributes and conditions influence recharge intensity and timing, which are among the main drivers of regional water table depth \cite{de2017global,maxwell2015high,moeck2020global}. Here, we calculated the long-term average values of climatic attributes, including precipitation, temperature, temperature of the month of January, rainfall intensity, maximum snow water equivalent (SWE), snow fraction, aridity index, and precipitation excess for the period between 1979-2009, which covers the time from which the bulk of real and proxy WTD observations are available. January temperature was used in addition to average temperature, as previous studies showed that winter temperature affects soil frost and, ultimately, WTD properties \cite{fan2013global}. To quantify long-term average precipitation, temperature, January temperature, rainfall intensity, and snow fraction, we used the EMDNA dataset \cite{tang2021emdna}, which provides daily precipitation and temperature at a resolution of 0.1 degrees. The long-term average snow fraction was calculated as the sum of snowfall (precipitation during days with below zero temperature) over the sum of precipitation. Rainfall intensity was calculated as the average rainfall intensity (mm/day) across days with over 1 mm of rain.

Long-term average values of potential evapotranspiration, actual evapotranspiration, and maximum snow water equivalent were estimated using the TerraClimate dataset \cite{abatzoglou2018terraclimate} at a resolution of 0.04 degrees. After converting the spatial resolution of all climate datasets to the reference 15 arc-sec cell size through geospatial resampling as discussed in subsection \ref{sec:data_resampling}, we combined gridded precipitation from EMDNA and potential evapotranspiration from TerraClimate to calculate the gridded continent-wide aridity index values as the ratio between potential evapotranspiration and precipitation \cite{addor2017camels}. Gridded precipitation excess was also calculated for North America as precipitation from EMDNA minus actual evapotranspiration from TerraClimate. The long-term average maximum snow water equivalent (SWE) was calculated as yearly maximum SWE from TerraClimate, averaged across all years. All climatic attributes were downscaled to our target resolution of 500 m (see Section \ref{sec:data_resampling}).

\subsubsection{Topography}

Conceptually, in most mountainous landscapes, the water table usually follows an extremely damped version of topography \cite{desbarats2002kriging,moreno2016modeling}. Indeed, in mountainous regions, water tables are often observed and simulated to be deep \cite{de2017global}. This is not only due to lateral groundwater flow seeking equilibrium with gravity-driven gradients but also due to decreased recharge at higher elevations \cite{moeck2020global}. Further, flat topography encourages discharge formations such as rivers, wetlands, and lakes, while mountainous terrains usually correspond to low-storage areas \cite{moeck2020global}. At the same time, precipitation increases with elevation while evapotranspiration tends to decrease. Hence, topography is one of the primary drivers of WTD, with the complex relationship between topography and groundwater recharge (and WTD) depending on how topography interacts with precipitation, evapotranspiration, and geological formations \cite{moeck2020global}. Machine learning may provide a strong foundation for exploring and simulating such a complicated relationship. Here, we used different topographical attributes which are known to control recharge and WTD. These attributes include elevation, slope, and topographic index. To cover the entire extent of the United States and Canada, we merged two elevation datasets. SRTM15+ \cite{tozer2019global} was used for zones below 60 degrees North, while GTOPO30 \cite{daac2004global} was used for a small portion of our study region with latitudes larger than 60 degrees. The former dataset is available at our target resolution (500 m), while GTOPO30 is a global digital elevation model (DEM) with a grid spacing of 30 arc seconds (approximately 1 km), which was converted to our target resolution using ArcGIS's spatial analysis toolbox, simply by splitting each 30 arcsecond pixel into four 15 arcsecond ones. Using the merged elevation dataset and GIS spatial analysis tools, terrain's slope was calculated in degrees. \citeA{marthews2015high} created a global 500 m resolution dataset of Topographic Index (TI), the index that was introduced by \citeA{beven1979physically}, to characterize the topographic control on the spatial distribution of water table. TI dataset was directly used in our model without any preprocessing.

\subsubsection{Geology}
\citeA{gleeson2014glimpse} developed global maps of bedrock permeability and porosity, two key properties for understanding groundwater hydraulics \cite{moeck2020global}. In our study, bedrock permeability and porosity were initially used as input data to the machine learning models to inform the geological characteristics of the shallow subsurface. However, our results showed close to zero importance for these two attributes, meaning that our machine learning models did not use the information of these two attributes to simulate the spatial pattern of static WTD at local to regional scales. 

Our machine learning models do not choose deep geology data as important factors potentially due to (1) low spatial variability of this data across the North America, (2) high uncertainties in deep geological data (as opposed to climate/topographic data, which are more certain) and (3) the coevolution of climate, soil, land cover and geology, leading to much of the information of deep geology already being encoded in other features. As will be introduced in the next section, incorporating hydraulic characteristics of deep soil could, to a large extent, emulate deeper subsurface hydraulic properties. In fact, due to the minimal variability of geology data, the information content that helps to predict static WTD seems to only be informed by correlated attributes such as (deep and shallow) soil hydraulic properties. This does not mean that bedrock hydraulic properties are unimportant, but it implies that the machine learning algorithm, which learns through the variability of predictors, opts to learn geologically relevant information/processes from highly variable and relevant soil hydraulic properties rather than almost invariant bedrock properties. Soil hydraulic properties (and even climate attributes) can be highly correlated with bedrock permeability in many environments, carrying important information about bedrock geology the currently available datasets of bedrock permeability cannot convey. We refer the readers to \citeA{heudorfer2024deep}, \citeA{ghosh2022robust}, and \citeA{li2022regionalization} for further details on how machine learning models may leverage the correlation/coevolution among attributes and learn the relevant information of a given uncertain attribute via other more certain attributes.

\subsubsection{Soil data}

While climate is the driver of overall catchment recharge, soil hydraulic properties control the partitioning between subsurface vertical infiltration (and recharge) and lateral flows \cite{maxwell2015high,moeck2020global}. Low soil hydraulic conductivity prevents water from entering the subsurface, encouraging the input water to exit the catchment via overland flow instead of contributing to soil moisture or aquifer recharge  \cite{maxwell2015high,moeck2020global}. \citeA{de2015high} argues that the soil saturated hydraulic conductivity is one of the most dominant controls of WTD, where higher conductivity lowers the water table and enhances regional flow. Soil thickness (and soil porosity) may also impact groundwater placement \cite{de2015high}.

The Soilgrids dataset \cite{poggio2021soilgrids} is a comprehensive global-scale information system providing high-resolution soil property maps for various depths and soil layers. The dataset covers a wide range of soil properties, including sand, silt, and clay fractions, organic carbon content, bulk density, depth to bedrock (or soil thickness), and more at a spatial resolution of approximately 250 meters. To characterize hydraulic properties of shallow portion of the earth surface, clay, silt, and sand fractions within the top 15 cm depth were aggregated to a resolution of 500 m. Similarly, the available data of clay, silt, and sand fractions below 1 m were aggregated to a resolution of 500 m to present the hydraulic properties of a relatively deeper portion of the Earth's surface material. These aggregated fractions at two different vertical positions were then utilized as input parameters for the machine learning groundwater models. Furthermore, the gridded depth to bedrock data was gathered from \citeA{dai2019global}, converted to 500 m resolution, and used as an additional soil-related input to our machine learning models.

\subsubsection{Land cover}

Vegetation plays a critical role in determining water input partitioning and ultimately groundwater recharge \cite{allen1998crop,moeck2020global}. Different types of vegetation have varying rates of transpiration, which is the process by which plants take up and release water into the atmosphere through the stomata on leaves. For example, a forested area with tall trees may have a higher rate of transpiration than a grassland, leading to a higher rate of evapotranspiration. This variation in transpiration and evaporation rates will affect the amount of water available to recharge the groundwater system, ultimately affecting the water table's depth. Changes in land use can impact recharge rates even if the climatic conditions remain the same, as noted by \citeA{defries2004land,minnig2018impact}. 

The Collection of Moderate Resolution Imaging Spectroradiometer (MODIS) Land Cover Dynamics Product, called MCD12Q, provides global land cover and phenology metrics at 500-meter spatial resolution. MCD12Q dataset includes annual land cover maps with 17 classes, including 11 natural vegetation classes, three human-altered classes, and three non-vegetated classes. In this study, MCD12Q's land cover map of the year 2000 was acquired and reclassed. Different forested land cover classes were combined to form one unique forested land cover class. Similarly, different shrubland classes and grassland classes were combined to form one unique shrubland class and grassland class, respectively. This led to a total of 7 land cover classes which were used as categorical inputs to the machine learning-based groundwater models. The classes selected for this analysis were chosen based on established findings on the relationships between land cover and groundwater recharge rate (and ultimately WTD) \cite{owuor2016groundwater,scanlon2005impact,mohan2018predicting,zhang2006effects}.

\subsection{Real Observations of Water Table Depth}
\label{sec:wtd_obs}
Water well observational information in the United States and Canada was compiled from national organizations and provincial databases. Water well observational information in the United States was downloaded from USGS's database on "Groundwater levels for the Nation" (\url{https://nwis.waterdata.usgs.gov/nwis/gwlevels}). In Canada, historic well reports are documented and made available for public use in most provinces. Details of provincial databases, data format, and web links for each province are provided in the Supplementary material document. A total of 9.6 million observations collected between 1852 and 2022 (with the majority collected between 1979-2009) were compiled at 905,371 observational wells across the USA and Canada. When multiple observations over time in a given well were present, the minimum value of WTD was taken to obtain one representative observation for the given well. This treatment enhances the chance of obtaining representative static WTD observations that are minimally influenced by pumping. Moreover, when converting point-based WTD observations to the target resolution of our study (i.e. 500 m pixels), there were instances, especially across flat agricultural lands, where two or more observational wells were situated within a single pixel. In such cases, the median WTD was designated as the pixel's representative observation of WTD. Figure \ref{fig:regionalObs} shows (in red) the 500 m resolution pixels with estimated representative observations of WTD across the Prairie Pothole region of North America and the Mississippi River Basin. Note that the mentioned processes in converting temporal point-scale observations to static pixel-scale WTD are inline with the previous large-scale (e.g., national or global) studies conducted to quantify static WTD. 

\subsection{Proxy Observations of Water Table Depth}

\subsubsection{Occurrence of surface water inundation connected to groundwater systems}
\label{sec:waterInundation}

Persistent water inundation on the Earth's surface is a strong indicator of hydraulic connection between surface water and the surrounding groundwater system \cite{de2017global,fan2013global,maxwell2015high,koch2019modelling}. Indeed, the static water table depth will often be zero or near zero at the interior or shoreline of persistent rivers, streams, wetlands, potholes, and lakes \cite{de2017global}. \citeA{pekel2016high} used more than three million Landsat satellite images to record the months (including summer) when water was inundated on the ground, between 1984 and 2015. They developed a probabilistic metric, Water Occurrence Percentage (WOP), which reflects the percentage of time at a given 30 m x 30 m pixel when water was present on the ground surface. The metric varies between 0 (no water) and 100 (permanent water), with values in between reflecting the intermittent presence of water. This milestone database was used in our study to develop two sets of WTD proxy observations. The first set of proxy observations reflects the permanently wet interior pixels of (typically small) surface water bodies and is used in our study as one line of independent evidence to evaluate and compare the performance of three machine learning and two physically-based simulations of WTD. The second set of proxy observations reflects the shorelines of surface water bodies of different sizes and is used in our study to train two (out of three) of our machine learning models. Note that there is no overlap between the first and second sets of pixels utilized for the first and second objectives. Consequently, the pixels that evaluate the performance of five machine learning and physically-based models are mutually exclusive from those used for training the two ML models. To preprocess the data, we first aggregated the 30 m resolution surface water inundation database to our target resolution (500 m). During this aggregation process, the representative WOP at each 500 m resolution pixel was calculated as the maximum WOP among all WOPs of the original 30 m resolution pixels. Similar to the original 30 m resolution database, the pixel's representative WOP at our target resolution ranged from 0 to 100.  

The first set of proxy data, aimed at delineating the permanently wet interior pixels of surface water bodies, was compiled by identifying pixels where surface water inundation was consistently detected over a specific occurrence threshold. Pixels experiencing water inundation for more than 95\% of the time (i.e. WOP$>$95\%) were delineated and merged. Then, we used the HydroLAKES dataset \cite{messager2016estimating}, which provides polygonal representations of water bodies, to select those pixels which were fully inside water body polygons. 7,920,475 pixels representing the interior of permanently wet surface water bodies were extracted and used to evaluate the two physically-based and three machine learning models' performances in locating pixels where WTD is expected to be zero (this will be further explained in Section \ref{sec:criteria}). HydroLAKES appears to have missed some water features with WOP$>$95\%. Notable examples are thousands of long and stretched water bodies, which may have been missed in surveying conducted to develop the HydroLAKES dataset, due to their narrow widths, restricted access, or other reasons. In such cases, merged pixels with WOP$>$95\% were considered as permanently wet surface water bodies' pixels. 
Regarding the selection of 95\% threshold of WOP, we have to clarify that visible and near-infrared range images used in \citeA{pekel2016high}, could have missed some inundated pixels in some months, because the light at these frequencies could be impeded by cloud cover, canopy, and ice. To account for this uncertainty, we incorporated an arbitrary error margin of 5\%, or the 95\% threshold of WOP, rather than using 100\% threshold of WOP to delineate permanently wet pixels. Opting for smaller error margins (e.g., 1-2\%) would misrepresent significant lakes, while larger than 5\% error margins would pose a risk of overestimating the permanently wet surface water bodies. Therefore, a WOP of 95\% was deemed the most appropriate threshold for this part of the study. Furthermore, we only relied on surface water bodies with surface areas between 30 and 80,000 hectares to ensure that a disproportionately large number of pixels were not selected from the interior of very large surface water bodies (e.g. the Great Lakes of North America). Indeed, excluding interior pixels of large water bodies enhanced the spatial uniformity of pixels, allowing for a fair evaluation and comparison of the models' performance. 

\begin{figure}[h!]
\captionsetup[subfigure]{aboveskip=-1pt}
     \centering
     \begin{subfigure}{0.6\textwidth}
         \centering
         \includegraphics[width=\textwidth]{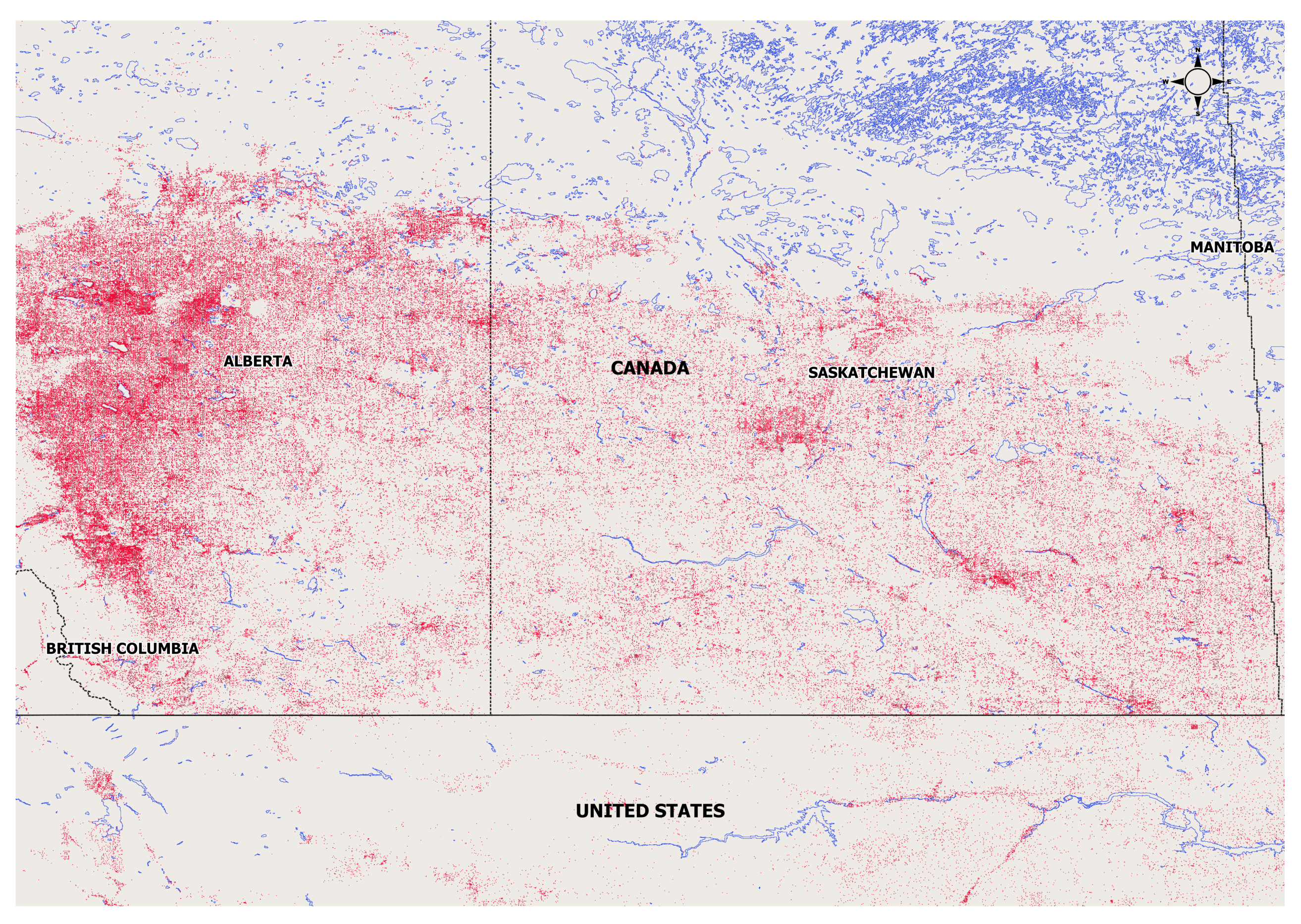}
         \caption{Prairie Pothole Region, Canada}
         \label{fig:PrairiePot_obs}

         \end{subfigure}
     \hfill
     \begin{subfigure}{0.352\textwidth}
         \centering
         \includegraphics[width=\textwidth]{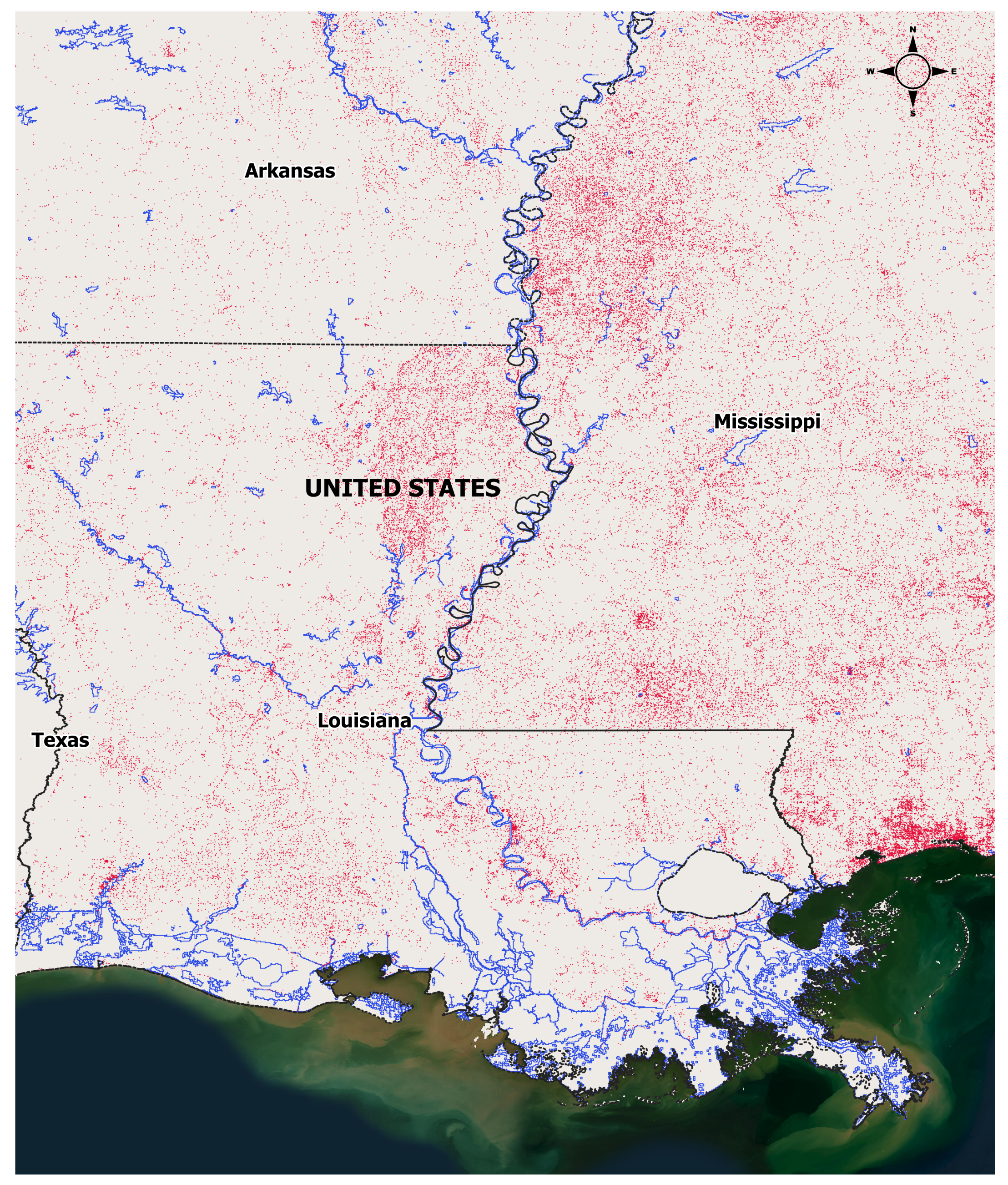}
         \caption{Mississippi, USA}
         \setcounter{subfigure}{3}
         \label{fig:Mississippi_obs}
          \end{subfigure}
     
        \caption{500 meter resolution pixels showing the locations of the available real observations of WTD (red) and that of the delineated shorelines of surface water bodies (blue), across the Prairie Pothole and Mississippi River Basin regions. Our real observations of WTD and the delineation of the shorelines of surface water bodies extend throughout the entire United States and Canada. But for the sake of simplicity of visualization, we only show these two selected regions with fairly dense observations and delineated shorelines.}
\label{fig:regionalObs}
\end{figure}

The second set of proxy data, aimed at delineating the semi-permanently wet shorelines of surface water bodies, was compiled again by pixel-based WOP values. Pixels meeting the criterion of semi-permanent to permanent inundation (WOP$>$75\%) were delineated and merged. Then, the peripheral, non-interior pixels around the groups of merged pixels were identified as the pixels of surface water bodies' shorelines. The selection of the 75\% threshold ensured the inclusion of areas beyond the interior of permanently wet surface water bodies that are still experiencing semi-persistent surface water inundation. Peripheral pixels with WOP$>$75\% indicate prolonged surface water inundation with potentially stable soil saturation, consistent with our definition of static WTD defined as the long-term non-transient WTD. Our preliminary analysis showed that choosing a larger threshold of WOP, such as 85-90\%, could have excluded a large portion of surface water bodies' shorelines delineated using other datasets (e.g., HydroLakes). On the other hand, opting for smaller thresholds (e.g., 60-50\%) risked including ephemerally inundated areas like wave run-up and overland runoff, which have no stable connection to surrounding groundwater systems. The 30 m resolution and the temporal information on surface water inundation provided by \citeA{pekel2016high} dataset facilitated the identification of the shorelines of numerous small and/or locally (and semi-permanently) inundated areas, which were not delineated by other available datasets such as HydroLAKES. In the end, 12,167,332 pixels were delineated as the pixels of shorelines of surface water bodies across the study region at a resolution of ~500 m. Proxy pixels also include river channels and flooded riparian zones where the watercourse is wide enough to result in WOP$>$75\%. These pixels were considered as the locations with static WTD equal to zero (i.e. the water table is typically to be at the land surface). They will be utilized as proxy observations of WTD to train two of our machine learning models (see Section \ref{sec:threeModels}). Figure \ref{fig:regionalObs} shows (in blue) the delineated locations of shorelines of surface water bodies across the Prairie Pothole region of North America and the Mississippi River Basin at 500 m resolution.

\subsubsection{Height Above Nearest Drainage}
\label{sec:HANDdata}
Digital Elevation Models (DEMs) inform the development of numerical descriptors relevant to identifying hydrological processes and WTD \cite{sivapalan1995scale}. One of WTD's most relevant DEM-based descriptors is the Height Above Nearest Drainage (HAND) \cite{nobre2011height,koch2019modelling}. Using a hydrologically coherent DEM, HAND values were calculated by (1) finding the drainage pixels using DEM-based upslope accumulated area. Drainage pixels are those with above some threshold value of upslope accumulated area. (2) The nearest drainage pixel was found for a given pixel, and their elevational differences were computed to obtain the HAND value for the given pixel. An extensive experimental campaign in northern Brazil by \citeA{nobre2011height} showed a strongly positive linkage between the static (non-transient) WTD and HAND values along steep mountainous regions. Indeed, WTD at a given pixel can strongly follow the surface elevation of its local drainage network. The global HAND dataset, originally calculated at a resolution of 3 arc-seconds \cite{yamazaki2019merit}, was utilized and resampled to match our study's reference resolution (500 m). The use of HAND data as additional proxy observations of WTD in our third machine learning model will be further explained in Section \ref{sec:threeModels}.

\subsection{Global-scale Physically-based Simulations of WTD}
\label{sec:PrevSim}

We compare our three machine learning-based simulations of WTD (see Section \ref{sec:threeModels}) with physically-based simulations developed by  \citeA{fan2013global}, and \citeA{de2015high} (data was gathered from \citeA{de2022model}). These two studies used historical data (1960-2010) to estimate static water table depth at the global scale. \citeA{fan2013global} used a simple 1 km gridded Darcy-based numerical scheme to simulate WTD. For the top 1 meter of soil, they used available datasets of hydraulic conductivity. They assumed an exponential decay of hydraulic conductivity for deeper geology and calibrated three parameters using water table depth observations in North American temperate zones. They also calibrated permafrost parameters using observations of wetland extent in Northern USA and Canada. \citeA{de2015high}'s global 11 km resolution simulation of WTD  was produced by coupling the land-surface model PCR-GLOBWB with the groundwater model MODFLOW. Each pixel in their coupled model has one canopy layer, two soil layers, and one groundwater layer. Groundwater flux is forced using temperature, potential evapotranspiration, and precipitation, while land cover, soil, and topography conditions are set using globally available datasets. Boundary conditions are set at large lakes and coastlines. Some hydrogeological parameters within the model of \citeA{de2015high} were chosen based on how well the simulated hydraulic heads fit the hydraulic heads estimated from piezometer observations. In the remainder of this paper, we refer to studies conducted by \citeA{de2015high} as de Graaf's simulation and \citeA{fan2013global} as Fan's simulation.

\subsection{Ecoregions}
\begin{figure}[h!]
     \centering
     \includegraphics[width=1\textwidth]{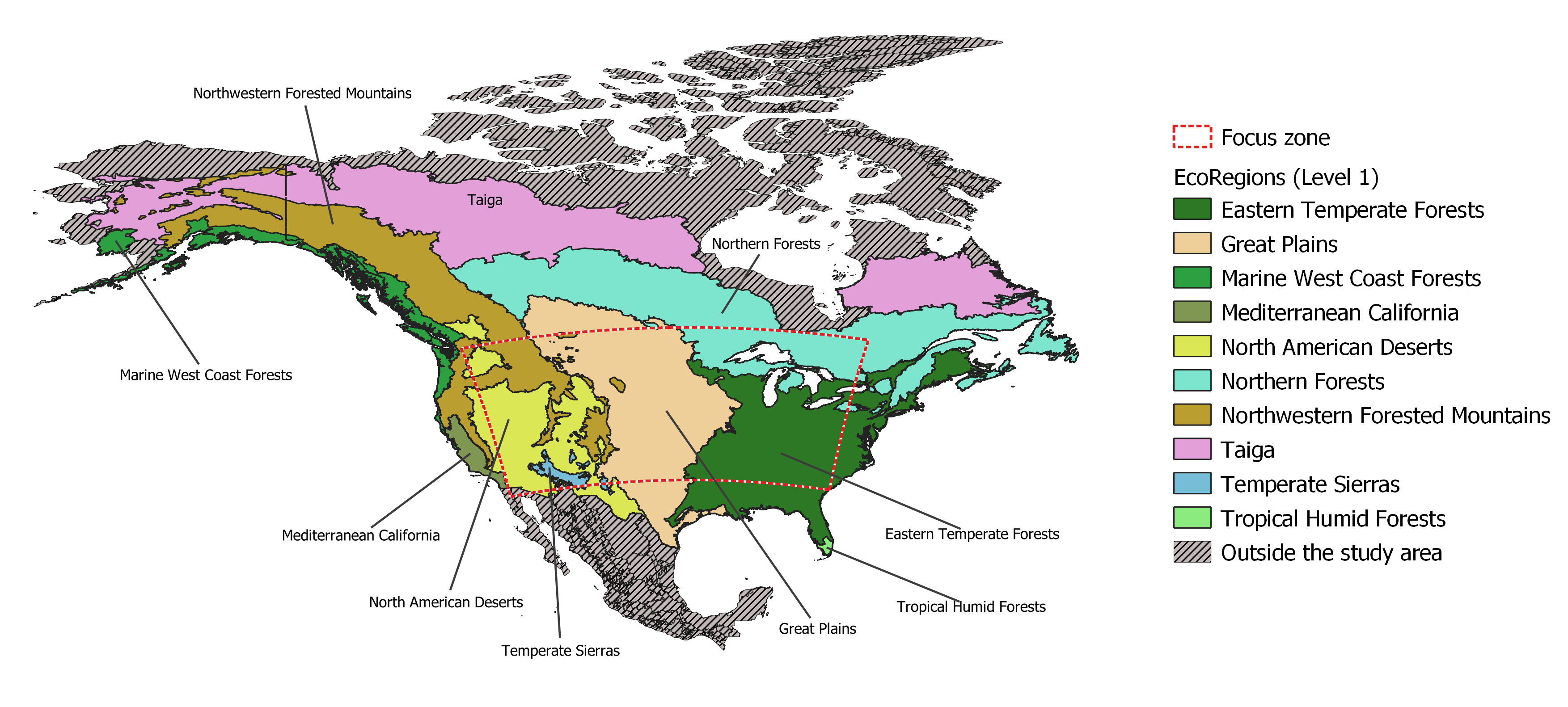}
     \caption{The locations of ten ecoregions along which we compare and evaluate the performance of the three machine learning-based and two physically-based WTD simulations. The dashed red arch-shaped zone referred to focus zone and was explained in Section \ref{sec:trainingvalidation}}
\label{fig:ecoRegions}
\end{figure}

Ecoregions are defined as areas of the Earth's surface that contain distinct assemblages of natural communities and environmental conditions. They are a way to divide the planet into relatively homogeneous areas based on similarities in climate, geology, topography, vegetation, and wildlife. The United States Environmental Protection Agency (EPA) divides North America into 15 different ecoregions based on the Omernik Ecoregion Framework \cite{omernik1987ecoregions}. The Omernik Ecoregion Framework is widely used by scientists and resource managers to understand and manage diverse natural ecosystems \cite{gallant1989regionalization}. These 15 ecoregions are: Arctic Cordillera, Tundra, Taiga, Marine West Coast Forest, Eastern Temperate Forest, Great Plains, North American Deserts, Mediterranean California, Southern Semi-Arid Highlands, Temperate Sierras, Tropical Dry Forest, Tropical Rainforest, Western Mountains, Valleys, and Coast Arctic Plains and Mountains. In each ecoregion, we compare and evaluate (using different lines of evidence) the performance of the three machine learning-based and two physically-based simulations of WTD. Note that the real observations of WTD were only available within ten ecoregions, and hence in the remainder of the paper, we only use the information of these ten ecoregions (Figure \ref{fig:ecoRegions}). Table \ref{tab:ecoreg_char} reports some statistics regarding these ten ecoregions' climatic, topographic, and soil characteristics. 

\begin{table}[h!]
 \caption{Mean and median climatic, topographic, and soil characteristics of ten ecoregions along which we evaluate model performance. The characteristics include long-term average Aridity index (-), Temperature ($^{\circ}C$), Elevation (m), Depth to Bedrock (DTB, cm), Shallow Clay (\%), and real observations of WTD (m). The ten ecoregions are (see \ref{fig:ecoRegions}): Taiga (3), Northern Forests (5), Northwestern Forested Mtns (6), Marine West Coast Forests (7), Eastern Temperate Forests (8), Great Plains (9), North American Deserts (10), Mediterranean California (11), Temperate Sierras (13), Tropical Humid Forests (15). Numbers in the parenthesis show the ecoregion class in the 15-class Omernik Ecoregion classification. Eco-region area is shown below the eco-region's class.} 
  \centering
  \begin{tabular}{lllllllll}             \\
     \hline
    \textbf{Region} & \textbf{Statistic} & \textbf{Aridity} & \textbf{Temperature} & \textbf{Elevation} & \textbf{DTB} & \textbf{Shallow Clay} & \textbf{WTD}\\
     \hline
    (3) & Mean  & 0.75 & -4.7 & 405 & 2331 & 17.1 & 4.12 \\
    $483km^2$ & Median & 0.74 & -4.65 & 331 & 2382 & 18.1 & 2.47 \\
    \hline
    (5) & Mean & 0.78 & 1.22 & 404.97 & 2039 & 18.2 & 6.75 \\
    $48140km^2$ & Median & 0.76 & 0.98 & 370 & 1835 & 18.4 & 4.57 \\
    \hline
    (6) & Mean & 0.93 & 0.89 & 1366 & 1641 & 15.9 & 19.1 \\
    $27406km^2$ & Median & 0.73 & 0.51 & 1233 & 1525 & 15.7 & 8.60 \\
    \hline
    (7) & Mean & 0.26 & 3.53 & 605 & 1723 & 12.4 & 7.92 \\
    $21301km^2$ & Median & 0.21 & 3.47 & 406 & 1628 & 11.9 & 4.57 \\
    \hline
    (8) & Mean & 0.88 & 13.15 & 205 & 1705 & 20.3 & 5.31 \\
    $122212km^2$ & Median & 0.89 & 13.3 & 181 & 1453 & 21.0 & 3.96 \\
    \hline
    (9) & Mean & 2.26 & 10 & 684 & 3908 & 28.1 & 17.0 \\
    $228570km^2$ & Median & 2.16 & 8.88 & 616 & 3755 & 28.3 & 9.14 \\
    \hline
    (10) & Mean & 6.44 & 13.4 & 1320 & 3659 & 19.5 & 27.7 \\
    $76231km^2$ & Median & 5.13 & 12.0 & 1398 & 2376 & 19.7 & 15.0 \\
    \hline
    (11) & Mean & 3.86 & 16.1 & 433.2 & 1282 & 21.6 & 19.6 \\
    $16278km^2$ & Median & 3.42 & 16.3 & 283 & 801 & 21.9 & 14.3 \\
    \hline
    (13) & Mean & 2.31 & 14.9 & 1935 & 529 & 23.7 & 40.7 \\
    $2858km^2$ & Median & 2.01 & 14.5 & 2019 & 188 & 23.1 & 17.9 \\
    \hline
    (15) & Mean & 1.14 & 25.1 & 98.0 & 361 & 35.0 & 1.46 \\
    $857km^2$ & Median & 1.16 & 25.4 & 29 & 248 & 36.8 & 1.19 \\
     \hline
  \end{tabular}
  \label{tab:ecoreg_char}
\end{table}

\section{Methods}

This paper aims to produce three sets of machine learning-based simulations of WTD (at 500 m resolution) in shallow unconfined aquifers across the USA and Canada. Section \ref{sec:data_filtering} explains our algorithm to filter out real WTD observations collected in confined aquifers. We then explain the resampling algorithm used to downscale coarse climatic input attributes to our target resolution (Section \ref{sec:data_resampling}). We explain the machine learning methodology, the physical constraints used to constrain the machine learning models, and the calibration procedures used to develop the three machine learning models (Section \ref{sec:MLmain}). We also explain how the three models were developed by gradually adding WTD proxy data to our training set (Section \ref{sec:threeModels}). The evaluation criteria used to evaluate and compare the performances of the three machine learning simulations and two physically-based simulations of WTD are discussed in Section \ref{sec:criteria}.

\subsection{Filtering of Real WTD Observations}
\label{sec:data_filtering}

This paper particularly focuses on the simulations of WTD within unconfined aquifers. Hence, we excluded the real well observations potentially collected within confined aquifers. Here, we leveraged the observations of WTD in the United States and British Columbia (Canada) flagged with information on confined versus unconfined status of the aquifers where the observations were collected. We explored whether well depth and geologic and climatic characteristics could explain the confined versus unconfined status of the observational data. We could then extrapolate this information to the other observational data where the confined versus unconfined status is unknown, but other characteristics are known. Since the ultimate aim was extrapolation, a powerful yet interpretable model was required. Hence, we employed a non-greedy decision tree algorithm, which provides accurate and interpretable extrapolation, using the \textit{evtree} R package \cite{grubinger2014evtree}. Inputs to the algorithm include climatic attributes, such as aridity index, precipitation, and snow fraction; geological attributes, such as depth to bedrock; topographical attributes, such as elevation; and well information, such as well depth. We found that aridity index (AI) and well depth were the strongest predictors of aquifer type. For dry regions (AI $>1.48$), wells at depths less than 241 meters are usually unconfined, while wells at depths greater than 241 meters are usually placed in confined aquifers. In more wet regions (AI $< 1.48$), wells at depths less than 19 meters are usually in unconfined aquifers, while greater than 19 meters usually indicate a confined region. The cross-validation metric showed that this simple yet interpretable model could identify unconfined real observations of WTD with over 85\% accuracy, while the overall accuracy in identifying the class of real observation (confined versus unconfined) was over 73\%. Although our overall accuracy is decent and our model is both simple and interpretable, we want to note that extrapolating to areas outside of the training set is always troublesome for data-driven models, but we believe this is the best currently available option. This model was applied to the entire study region to label real WTD observations (and their corresponding wells) as confined versus unconfined. Finally, we excluded all observations labelled as confined. In the end, this led to 541,418 pixels (at 500 m resolution) of real observations of WTD. Other criteria for aggregating point-scale to pixel-scale observations can be found in Section \ref{sec:wtd_obs}.

\subsection{Resampling of Climate Data}
\label{sec:data_resampling}
This study aimed to estimate water table depths at a resolution of 15 arcseconds (500 m at the equator). However, the continental-scale climatic input variables explained in Section \ref{sec:inputdata_climate} were only available at coarser resolutions, such as 360 arcseconds. The coarse-resolution climate datasets in GIS raster formats were resampled to the 15-arcsecond target resolution using the Geographically Weighted Regression method (GWR). In this technique, the independent variable of the down-scaling was elevation. We used the GWR for Raster Down-scaling in the SAGA software package (Spatial and Geostatistics menu).

\subsection{Machine Learning Modeling of WTD}
\label{sec:MLmain}

\subsubsection{Machine learning algorithm: XGBoost}
\label{sec:xgboost}

Decision trees can be simple yet powerful and interpretable models of complex data. Methods that combine a large number of decision trees into an ensemble exhibit strong predictive capabilities and computational efficiency on tabular data \cite{shwartz2022tabular}. Tree-based ensemble methods include random forests \cite{breiman2001random} and gradient-boosted decision trees such as XGBoost \cite{chen2016xgboost}, CatBoost \cite{prokhorenkova2018catboost}, and LightGBM \cite{ke2017lightgbm}. While random forests is the most popular tree-based ensemble method in hydrology \cite{tyralis2019brief}, our preliminary results showed a stronger performance of XGBoost on our dataset. Other studies also showed the strong performance of XGBoost in the simulation of groundwater systems \cite{osman2021extreme}. XGBoost, the machine learning algorithm used in our paper to simulate WTD, is centred around the sequential optimization of shallow decision trees \cite{james2013introduction}. A single shallow decision tree is a \textit{weak} learner as it captures the broad relationships in the data with a minimal chance of overfitting. The sequential nature of the algorithm allows XGBoost to learn slowly while correcting itself during the learning process. This is captured by iteratively regressing the residuals of the previously built set of shallow trees. The learning rate and level of stochasticity can be controlled via various hyperparameters. 

Building an XGBoost algorithm requires robust calibration and validation of regular XGBoost parameters and optimization of hyperparameters \cite{probst2019tunability,bilolikar2023out}. In developing each model, the (real and proxy) WTD observations are split into a training, validation, and test set. The regular XGBoost parameters are calibrated using the training set, and hyperparameters are calibrated against the validation set.  20 iterations of random search were used to select hyperparameters based on the model's performance ($R^2$) on the validation set. Once hyperparameters are chosen, the model is retrained on both the training and validation sets and tested on the test set. We report the performance of the three machine learning models against (real and proxy) WTD observations of the test set. The final simulation of each machine learning model is then made based on the full model trained on all data.

\subsubsection{Training, validating and testing sets}
\label{sec:trainingvalidation}

For training, validation, and testing our machine learning models, we divided the observations into two groups. The first group includes pixels within an arc-shaped focus zone across the United States (Figure \ref{fig:ecoRegions} shows this zone in red), and the second group includes the remainder of the pixels across our entire study domain (USA and Canada). The arc-shaped focus area encompasses a large portion of the USA and small parts of Canada, including the most of the Eastern Temperate Forests, Great Plains, North American Deserts, and significant portions of the Northern Forests and Northwestern Forested Mountains, including the USA's Rockies. It excludes the Eastern and Western coasts, large portions of Florida, Texas, California, Oregon, and most parts of Canada.  This zone has been used as the study domain in large-scale WTD simulations by \citeA{maxwell2016connections}, \citeA{maxwell2015high}, \citeA{ma2024water}, and \citeA{yang2023high}. This focus zone contains a large portion (about one-third) of our real observations of WTD and includes USA's most important groundwater-dependent ecosystems and resources development landscapes while excluding some highly exploited aquifers (e.g., California's Central Valley Aquifer). Hence, accurate prediction (and extrapolation) of static WTD is both plausible (given dense WTD data availability) and necessary within the focus zone. In our study, the training set includes 1/3 (randomly selected) of the real and proxy observations within the focus-zone domain and 2/3 of the (real and proxy) observations from the rest of study region. The validation set includes 1/3 of the (real and proxy) observations within the focus-zone + 1/3 of observations from the rest of the study region. The test set includes the remaining 1/3 of (real and proxy) observations within the focus-zone domain. 

\subsubsection{Implementing physical constraints}
\label{sec:montone}
Groundwater systems are inherently complex. This complexity is influenced by various factors, including climate, land use, topography and geology, and how they interact together, which need to be included in numerical models to effectively represent groundwater dynamics \cite{ojha2015current}. While machine learning techniques have the potential to capture these complexities to some extent, their flexibility can lead to models that may become unwieldy or difficult to interpret. To avoid this, some constraints should be forced on the internal relationships of machine learning models to ensure logical consistency in their predictions \cite{bierkens2019non}. Such constraints can also enhance the generalization and extrapolation capabilities of machine learning models \cite{janssen2021hydrologic}.

One standard course of action in machine learning modeling is to use domain knowledge to constrain the model via physically meaningful monotonic relations among input variables and the response \cite{fallah2012learning,ben2009adding,bartley2019enhanced,cano2019monotonic}. A monotone increasing (decreasing) relationship between a variable $x_j$ and the response $y$ is defined as follows: given an increase in $x_j$, and that all other variables are held constant, the response should not decrease (increase) in value \cite{bartley2019enhanced,fallah2012learning}. Utilizing monotonic constraints can decrease the size of the search space of \textit{good} models, ensure sensible model behavior and ultimately increase generalizability \cite{gutierrez2016current,bartley2019enhanced,cano2019monotonic}. While monotonicity is a strong assumption, it is clearly applicable across many areas of science \cite{liu2020certified}. One prominent example of this, as suggested by \citeA{fallah2012learning}, is that doctors would not trust a model where increased cigarette consumption leads to a predicted decrease in lung cancer risk. Likewise, groundwater hydrologists should not trust a model in which increased precipitation from one pixel to the other, while keeping all other variables constant, leads to increased WTD predictions. XGBoost can easily incorporate such physically meaningful monotonic constraints \cite{dong2022data,bartley2019enhanced,ovchinnik2019monotonicity,yang2021reliability}. In doing so, XGBoost rejects any split that violates the direction of monotonicity \cite{dong2022data}. Further, the mean values of a split are passed down to its children such that monotonicity cannot be violated further down in the tree \cite{dong2022data,bartley2019enhanced}. This simple framework allows for guaranteed global monotonicity while not dramatically increasing computational cost \cite{bartley2019enhanced,ovchinnik2019monotonicity}. While this approach has been used extensively in other areas of science, we are not aware of any previous applications of monotonic constraints in groundwater hydrology. 

Groundwater levels are influenced by various terrestrial characteristics and  environmental factors that govern recharge processes and groundwater dynamics \cite{delin2007comparison, jan2007effect, nolan2007factors}.  A comprehensive overview of the climatic, geological and topographical factors affecting groundwater level and recharge rate is provided by \citeA{moeck2020global}. The location of the water table is a function of net groundwater recharge and how fast the system retains or laterally drains groundwater \cite{maxwell2015high}. For instance, a higher aridity index, all else being equal, cannot lead to a higher recharge rate \cite{brutsaert1979advection, moeck2020global}. We know from the simple mass-balance equation that spatial increases in long-term average precipitation and decreases in evapotranspiration, all else being equal, cannot lead to a decrease in recharge rate and groundwater storage. Similarly, a larger topographic index at a given pixel, all else being equal, implies a larger amount of water flowing toward the pixel and should not allow for deeper WTD. A larger sand fraction, all else being equal, suggests faster vertical/lateral drainage of groundwater, leading to deeper WTD. Conversely, a larger clay fraction suggests a slower groundwater drainage, leading to shallower WTD. Physically-based models also satisfy these monotonic relationships if they converge and meet the constraints imposed by mass balance and momentum equations. 

Building on the above scientific rationals, we imposed the following constraints to all three set-ups of machine learning models developed in our paper: (1) WTD must monotonically increase with a larger aridity index, and with larger deep/shallow sand fractions, and (2) WTD must monotonically decrease with precipitation, precipitation excess, deep/shallow clay fractions, and topographic index. Again, it is important to note that within each monotonic constraint, we technically assume that all other variables remain unchanged. Within the XGBoost model, this implies that between two pixels with identical geology, topography and land cover, the one with a larger aridity can not have shallower simulated WTD. Alternatively, between two pixels with identical climate, land cover, and topography, the one with a larger deep/shallow sand fraction and smaller deep/shallow clay fraction can not show shallower WTD. While such constraints can be precisely imposed on machine learning models, the situation in which all climatic and biophysical factors are identical except for one factor may not always happen in reality. However, the imposed constraints intuitively follow basic principles from groundwater hydrology and, as explained before, are the ones that physically-based models should also satisfy.  Despite their simplicity, these constraints can mitigate the risk of equifinality of machine learning models by significantly reducing the size of the search space of \textit{good} model. 

With further advances in the algorithms of machine learning models and with the continued identification of increasingly complex (nonlinear) relationships among hydro-geological factors, future works should be able to train more robust physically-constrained machine learning-based groundwater models. Note that the (nonlinear) functional relationships among hydro-geological factors in shaping water table depth can be derived using fine-resolution physically-based models or advanced causal inference methodologies. While the former is a plausible near-future target, advanced causal inference methodologies able to identify complex nonlinear relationships from uncertain observational data have not been developed yet. Until then, the simple yet intuitive constraints used in our paper can be considered the first attempt in groundwater hydrology to improve the physical realism of machine learning models. Note that during our initial experimentation, we found that the model fit and test set $R^2$ decreased slightly due to adding the above constraints. \citeA{ben2009adding} and \citeA{bartley2019enhanced} reported a similar condition and finding. Indeed, the unconstrained XGBoost model may overfit to the noise and biases in the observational data; obviously these noise and biases, as will be discussed in Section \ref{sec:uncertain_data}, do not obey any expert knowledge-based monotonic constraints \cite{ben2009adding}. 

Note that to evaluate the need for the use of constraints in our models, we conducted metamorphic (or stress) testing, which is a widely-used analysis to evaluate the presence of poor extrapolation behavior in machine learning models \cite{yang2021reliability,reichert2024metamorphic,wi2024need,wi2022assessing}. In this test, we compare the reliability of our ML model with and without physical constraints, under the hypothetical scenario of increased temperature (+4 degrees) and decreased precipitation (-20\%). A reliable model should not show shallower simulated WTD under these much drier conditions. The ML model without physical constraints showed shallower simulated WTD in about half of the study region pixels, while in the model with constraints, none of the pixels can become shallower with increased temperature and decreased precipitation. Considering this preliminary stress testing analysis (results not shown), it is clear that there is a need to incorporate physical constraints to help the model better understand the relationships between physical/climatic attributes and WTD.

\subsubsection{Three machine learning model setups}
\label{sec:threeModels}
Three different sets of (real and proxy) observations of WTD were used to optimize XGBoost's (hyper)parameters, leading to the development of three machine learning models and WTD simulations. In the first model (V1), we only used 541,418 pixels of real observations of WTD from unconfined aquifers, with each real observation receiving a weight equal to 1. As we will see in the results section, this model could accurately simulate WTD along the observational locations while  missing excessively shallow WTD close to permanent surface water bodies. Hence, we introduced the second model (V2), in which 12,167,332 pixels of WTD proxy data along the semi-permanently wet shorelines of surface water bodies were added to the real observations of WTD, in order to optimize the (hyper)parameters of the XGBoost algorithm. WTD was assumed to be zero at these pixels. These WTD proxy data received a weight equal to the magnitude of water occurrence probability (WOP) at the given pixel, which should be larger than 75\%. Indeed, more persistent water inundations are more likely to have a zero WTD. See Section \ref{sec:waterInundation} for more details on delineating the semi-permanently wet shorelines of surface water bodies and the quantification of WOP. 

As we will see in the results section, despite very accurate performance of V2 along and close to (real and proxy) observational locations, V2 led to excessively shallow WTD along some high-elevation steep mountainous landscapes. This contradicts the findings of local-scale studies, which have consistently observed a deep WTD along high-elevation mountainous regions \cite{manning2013links,smerdon2009approach,ofterdinger2014hydraulic,somers2019groundwater}. Indeed, due to a lack of real and proxy observations along high-elevation steep mountainous landscapes, V2 may not properly \textit{learn} groundwater processes occurring along these landscapes. Height above the nearest water body (HAND) was shown to approximate WTD \cite{koch2019modelling,nobre2011height}, especially in dry regions with mountainous topography \cite{gleeson2011classifying,desbarats2002kriging}. Hence, HAND values in the mountainous regions of North America may offer a fair approximation of WTD in these regions. Thus, we introduced the third model (V3), in which 2\%\ of pixels, with HAND values larger than 30 meters (a total of 766,814 pixels), were randomly sampled and added to the (real and proxy) observations used in V2 (see Section \ref{sec:HANDdata} for more details on HAND data). Along these pixels, WTDs were assumed to be equal to HAND values. Note that the 30-meter threshold represents the third quartile of HAND values, corresponding to the steepest and highest lands across the study region. HAND data is a more uncertain proxy of WTD than the wet shorelines of surface water bodies, thus, a smaller weight of 0.75 was used for HAND data in V3 model development and optimization.  

\subsubsection{Interpretation of machine learning models}
\label{sec:interpMethods}
The same climatic and physical input variables, and the same physical constraints, were used for all three models (input variables were explained in Section \ref{sec:inputdata}). To interpret what the machine learning algorithm learnt, we can quantify and rank the extent to which each XGBoost model relies on each input variable. To achieve this for each model, we computed each input variable's gain feature importance. The split-level gain metric of a given input variable was computed as the gain in accuracy after each split belonging to the input variable. The split-level gains were then averaged across all splits in the ensemble of trees to form the final gain metric of each input variable for each model \cite{chen2019package}. Such a metric aims to reflect the structure that the machine learning model learnt during model development (or optimization) and can be used to infer the extent to which such structure follows available scientific literature relevant to WTD prediction.

\subsection{Model Evaluation Criteria}
\label{sec:criteria}

We compared the performances of our three machine learning-based and the two physically-based simulations of WTD within individual ecoregions and across the entire study domain. The \textit{unseen} data against which the models' performances are evaluated include (1) \textit{unseen} real WTD observations (OBS) and (2) proxy data on the permanently wet interior pixels of surface water bodies (WB), wherein we expect WTD to be zero (see Section \ref{sec:waterInundation} for more details on the delineation of these pixels). As explained in Section \ref{sec:trainingvalidation}, across the entire study region, unseen or test set real WTD observations for the three machine learning simulations include randomly selected one third of available real WTD observations within arc-shaped focus zone (Figure \ref{fig:ecoRegions} shows this zone), which were not used in machine learning models development. Within each ecoregion, unseen real WTD observations were identified through 10-fold cross-validation. Proxy data from permanently wet interior pixels of surface water bodies were not used in machine learning model development and were technically unseen for the three machine learning simulations. The two physically-based simulations, to some extent, relied on real observations of WTD and/or the location of surface water bodies to obtain more optimal model parameters (see Section \ref{sec:PrevSim}). However, to be fair to these simulations, in our paper, we assume that all real WTD observations and the interior pixels of surface water bodies were unseen by these physically-based models. Hence, for the two physically-based simulations, all real and proxy WTD observations were considered unseen across the entire study region and each ecoregion. 

For all machine learning and physically-based simulations, at pixels where unseen real observations were located, we computed the mean absolute error (MAE) between simulations and real observations of WTD (MAE-OBS). Similarly, along the interior pixels of water bodies wherein zero WTD is expected, we computed the mean absolute error or difference from zero (MAE-WB). The authors of some of physically-based simulations of WTD usually argue that their simulations of WTD are biased replicas (and not accurate simulations) of the observed WTD, as some of physically-based models do not account for human intervention \cite{fan2013global,maxwell2015high,yang2024concn,yang2023high,rahman2023simulating,de2015high}. Human intervention due to pumping could significantly bias the observations of WTD. Hence, for all 5 different simulations, we also estimated the correlation between simulations and real observations of WTD (Corr-OBS) along pixels where unseen real observations were located. In doing so, we evaluated how much each model captures the spatial variability of WTD without harshly punishing physically-based simulations for possible pumping biases. In addition to comparing the performances of machine learning and physically-based models model simulations against unseen real and proxy observations of WTD, we explored the pixel-by-pixel statistical association among the five simulations, across the study region. We do this by computing the correlation between each pair of simulations and visualizing it as a correlation matrix. Such inter-model correlational analysis can reveal similarities between two simulations, while helping us to determine output trustworthiness. Although the similarity of two model outputs is not a direct measure of correctness, two independent labs coming to the same conclusions is a good measure of replicability, and two models trained with different subsets of data leading to similar outputs shows stability. Stability and replicability are indirect measures of trustworthiness and are a lower bound on the requirements of scientific discovery/usefulness \cite{yu2013stability,fonio2012measuring,barnett2016validation,open2015estimating}.

We also evaluate the residual distribution of all five models across different quartiles of 6 input attributes, namely aridity index, elevation, precipitation, depth to bedrock, shallow sand fraction, and deep silt fraction. This analysis can reveal how bias and variance in WTD estimation change across models at different levels of climate, soil, and topographic attributes.

\begin{figure}[h!]
     \centering
\includegraphics[width=0.99\textwidth]{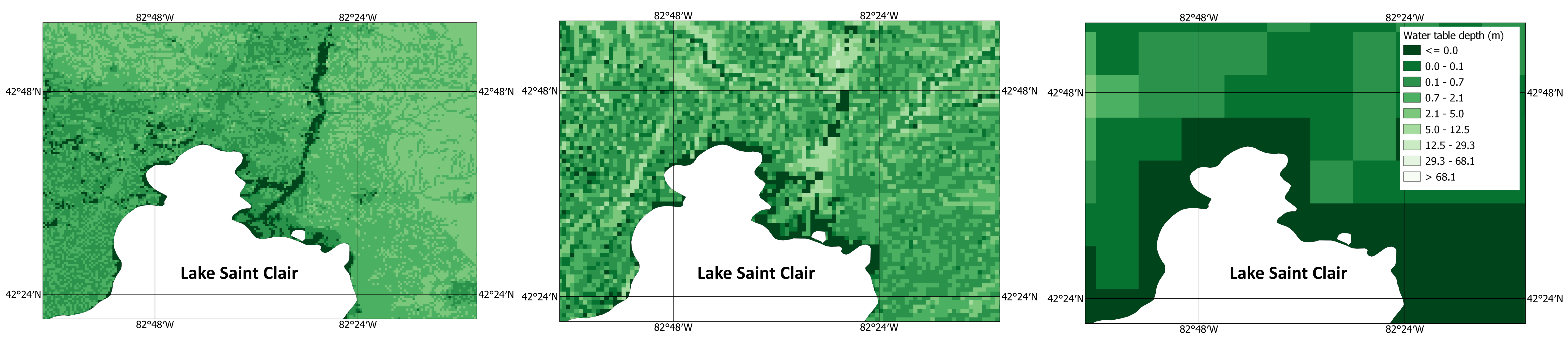}

     \caption{Simulations of Degraaf's (right), Fan's (middle) and our V3 (left) after the alignment process in the lowland area north of Lake Saint Clair. All three simulations identify the location of Lake Saint Clair and it's shorelines (with close to zero WTD) at similar locations. Shorelines are simulated at finer resolution using our V3 model than Fan's and Degraaf's simulation, due to differences in grid resolution.}
\label{fig:CellAlign}
\end{figure}

In order to accurately integrate different datasets of input variables and compare different simulations, we should ensure that all pixels and geographical points of datasets and simulations are perfectly aligned. Different datasets can sometimes be slightly offset due to differences in the underlying reference systems or grid resolution. In this study, we adjusted the alignment of all input datasets to machine learning models as well as two physically-based simulations, in a manner similar to \citeA{tootchi2019multi}. Pixel locations and alignments along the Pacific and Atlantic Oceans shorelines as depicted in the SRTM+ dataset (geographic coordinate system: World Geodetic System 84 - WGS 84 - EPSG:4326) were considered as the reference spatial pixels and system. Then, if needed, we aligned the oceans' shoreline locations in all datasets of input variables and two physically-based simulations.  For instance, Figure \ref{fig:CellAlign} depicts that after re-alignment, our machine-learning simulation and two physically-based simulations similarly identified the location of a small lake as well as its shoreline (pixels with close to zero WTD).

\section{Results}
\label{sec:Results}

\subsection{Description and Comparison of the Spatial Patterns of WTD Simulations} 
\label{sec:general_wtd_results}

\begin{figure}[h!]
\captionsetup[subfigure]{aboveskip=-1pt}
     \centering
     \begin{subfigure}{0.401\textwidth}
         \centering
         \includegraphics[width=\textwidth]{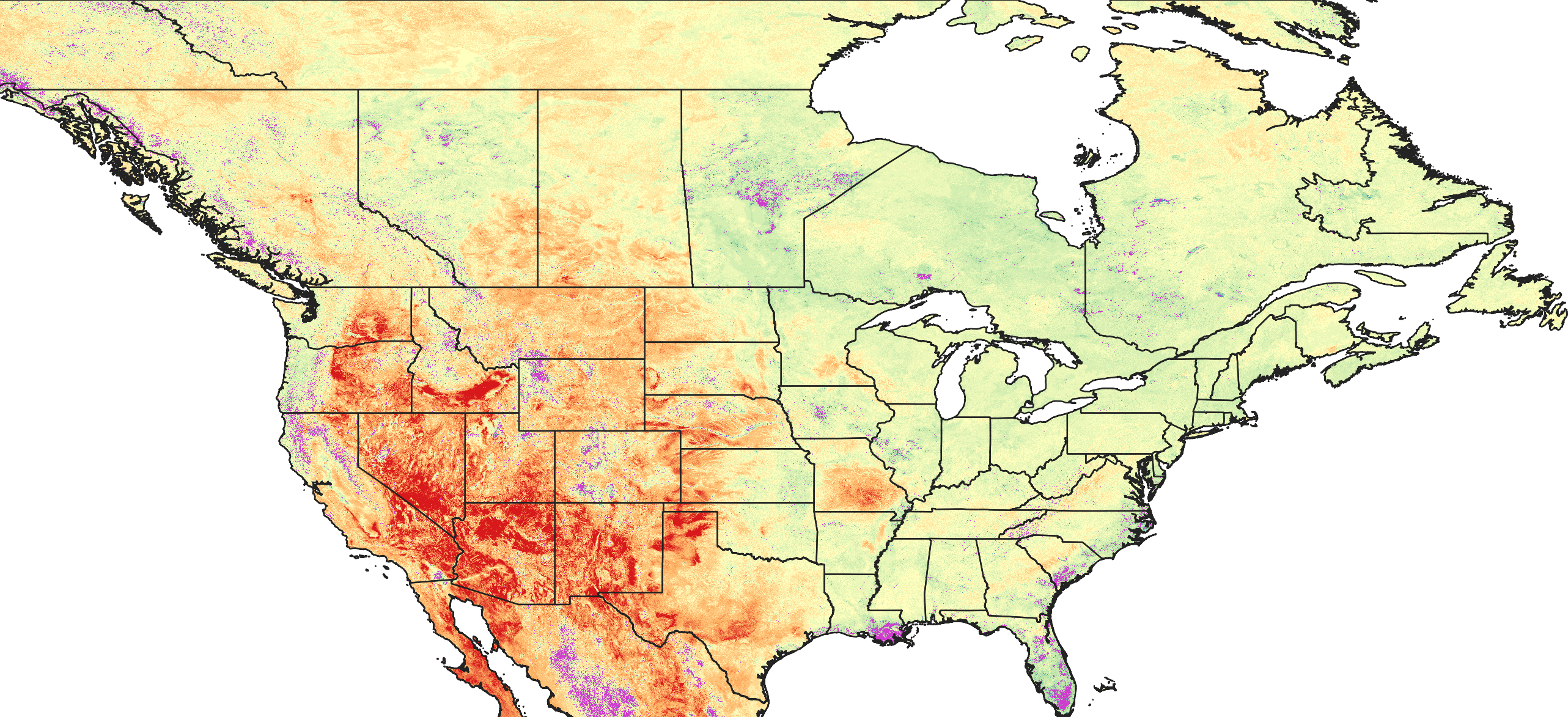}
         \caption{XGBoost - V1}
         \label{fig:simV1}
         \setcounter{subfigure}{1}
         \includegraphics[width=\textwidth]{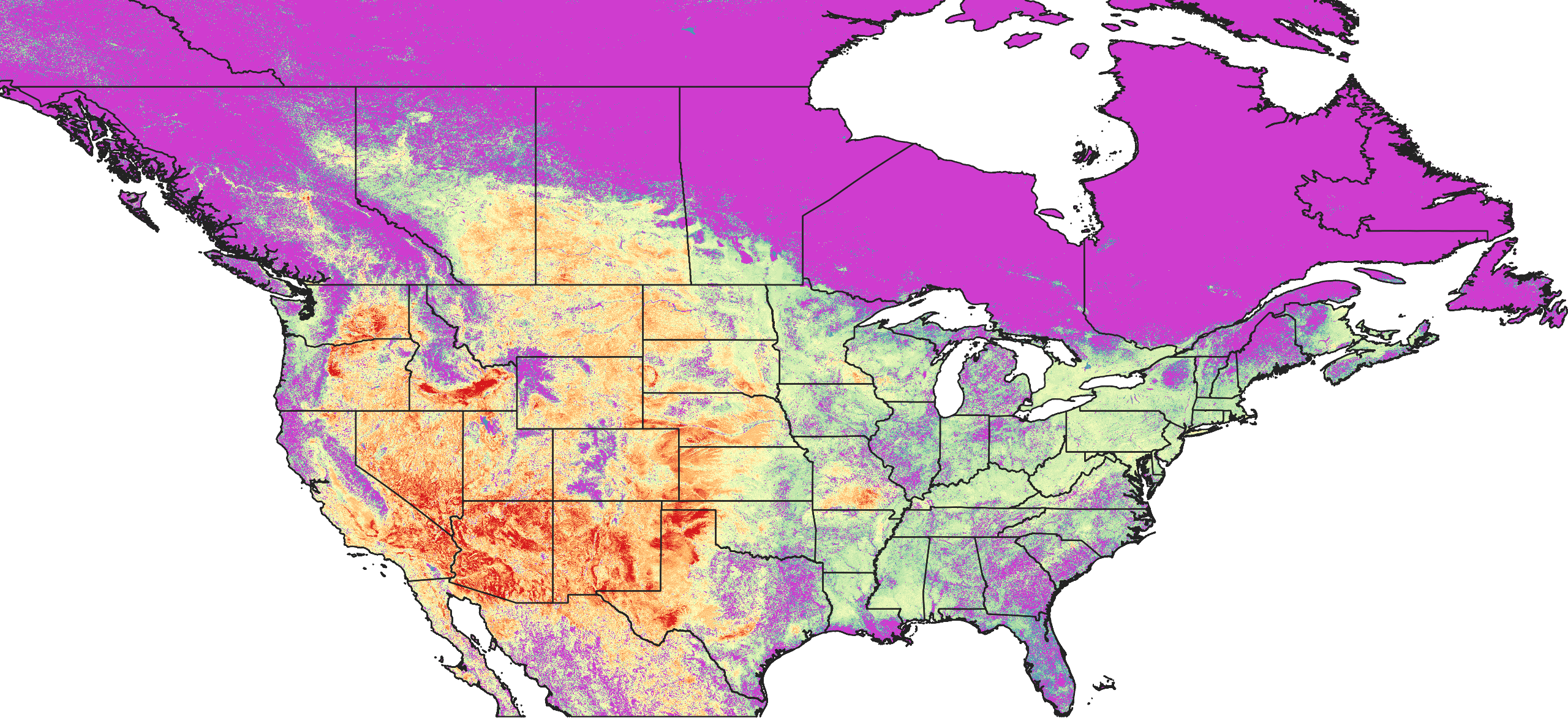}
         \caption{XGBoost - V2}
         \label{fig:simV2}
         \setcounter{subfigure}{2}
         \includegraphics[width=\textwidth]{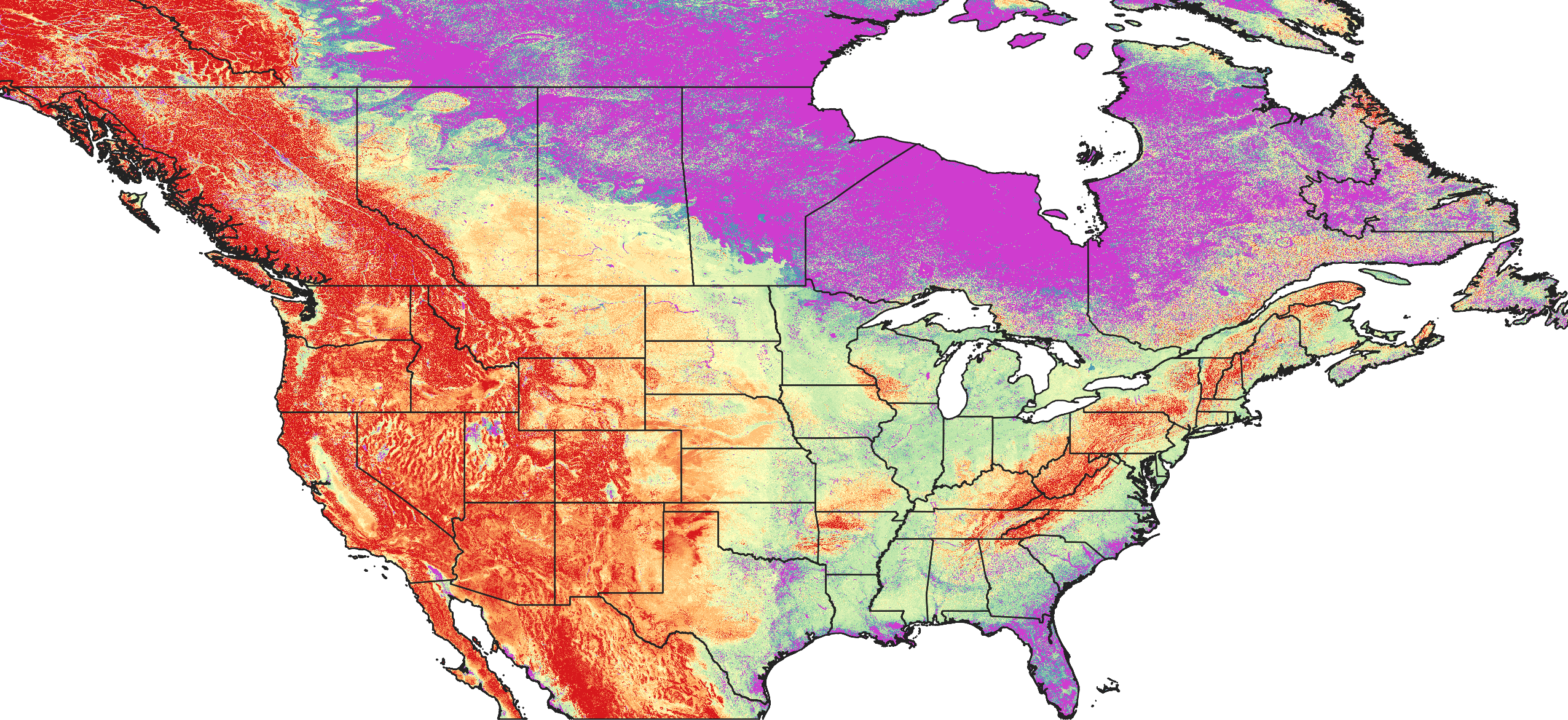}
         \caption{XGBoost - V3}
         \label{fig:simV3}
         
         \end{subfigure}
     \hfill
     \begin{subfigure}{0.401\textwidth}
         \centering
         \includegraphics[width=\textwidth]{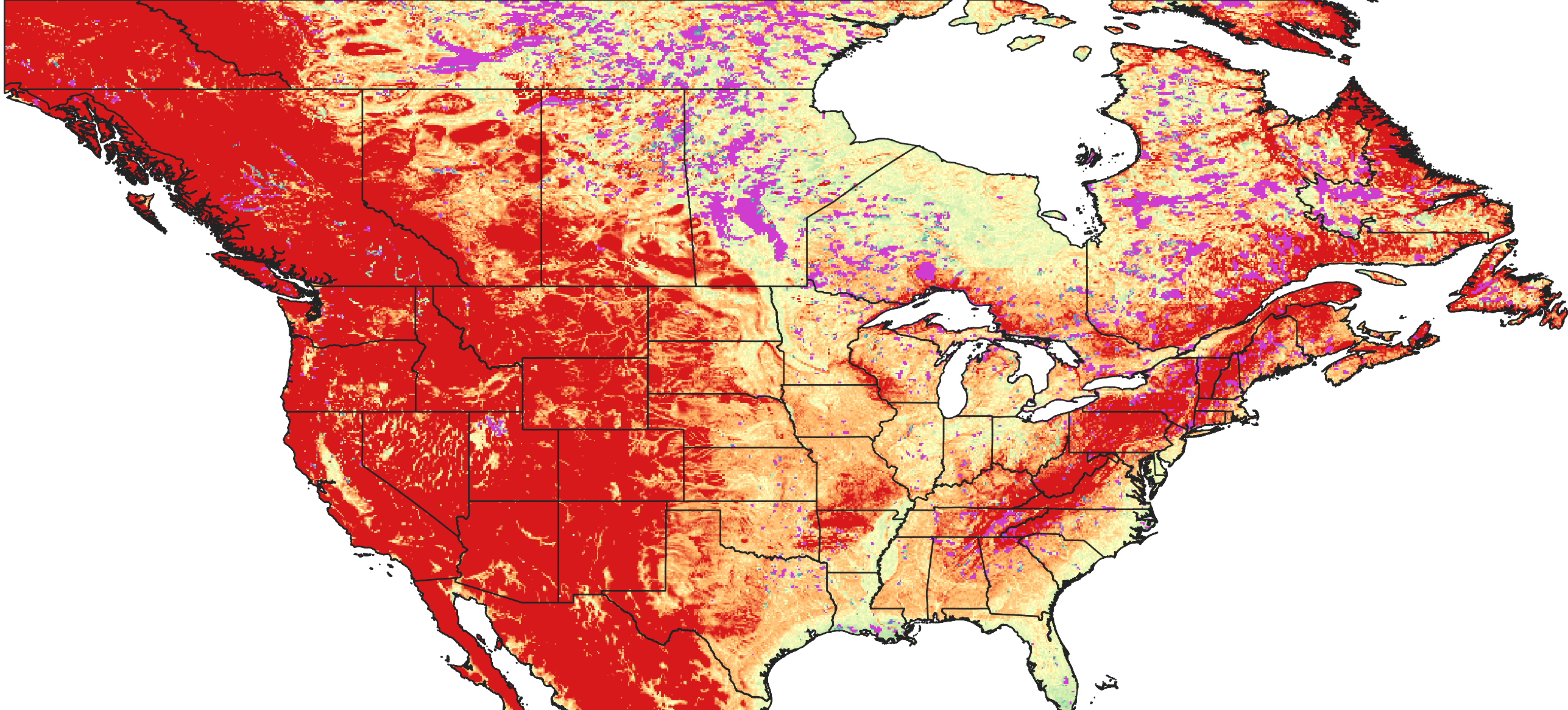}
         \caption{de Graaf's Simulation}
         \setcounter{subfigure}{4}
         \label{fig:simWTDDegraaf}
         \includegraphics[width=\textwidth]{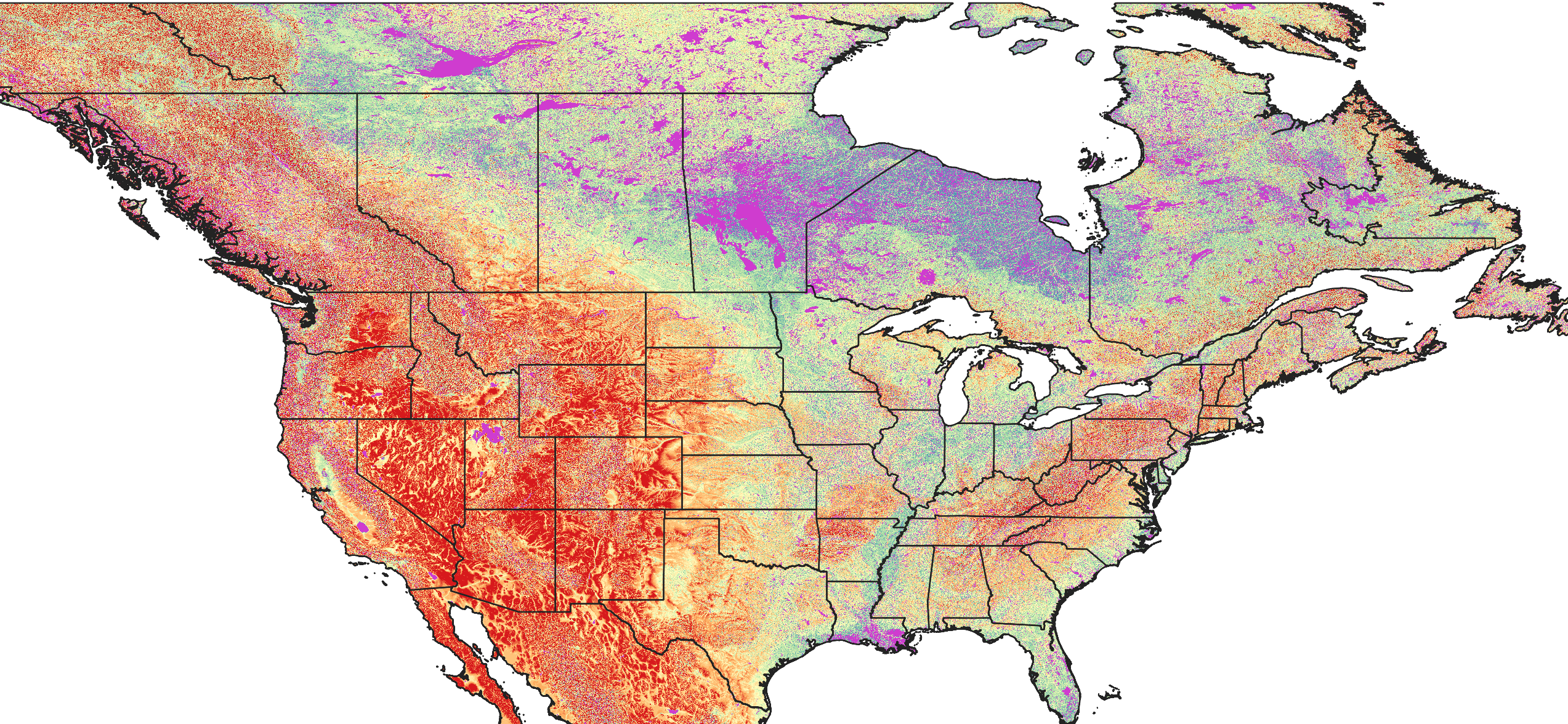}
         \caption{Fan's Simulation}
         \label{fig:simWTDFan}
         \setcounter{subfigure}{5}
          \end{subfigure}
     \hfill
     \begin{subfigure}{0.181\textwidth}
         \centering
         \includegraphics[width=\textwidth]{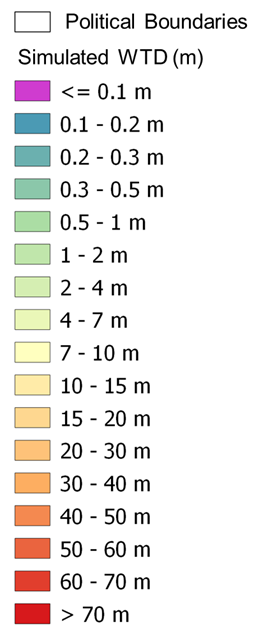}
         \caption*{ }
     \end{subfigure}
     
        \caption{Five different simulations of water table depths across the US and Canada using machine learning models (a-c) and physically-based models (d-e).}
\label{fig:SimulatedWTD}
\end{figure}

Overall, the three machine learning simulations consistently show shallower WTD on the eastern portion of the continent of North America than the western portion (\Cref{fig:simV1,fig:simV2,fig:simV3}). However, the WTD simulations of the three machine learning models exhibit significant differences at local scales. V1 could not reasonably predict WTD close to (and within) lakes and rivers, particularly along lakes in Canada's North and Prairie Pothole Region, where WTD was predicted to be around 10m. In contrast, V2 accurately predicts that WTD should be near zero close to (and within) lake and river locations. However, V2 also leads to excessively shallow (less than 0.1 m) WTD along a large portion of Canada (with the exception of the Canadian Prairies, South-eastern Canada and central British Columbia), while V1 is more conservative with predictions of 1-5 m in such locations. Regardless, the overall spatial patterns in V1 and V2 are similar, and these two simulations are highly spatially correlated (0.78), while V3 has a negligible spatial correlation with two other XGBoost models (0.11-0.16) (\Cref{{fig:CorPlot}}).

One notable distinction across the three machine learning models occurs along the Canadian Rocky Mountains, where V2 shows very shallow groundwater simulations, while V3 estimates the depth to be deeper than 50m along high-elevation steep lands, and V1 predicts a WTD of up to 10m. Furthermore, across the Canadian Rocky Mountains, WTD is highly spatially variable in V3, following the large spatial variations in elevation and slope, while V1 and V2 predict fairly spatially uniform WTD. Unlike V1 and V2, V3 follows the general patterns and magnitudes of Fan and de Graaf's simulations across the Canadian Rocky Mountains. Similarly, across the USA's Appalachia, V3 clearly distinguishes between the Appalachian Mountains of the eastern USA and the rest of the eastern USA. While V1 (V2) shows a fairly uniform distribution of WTDs in this region with no clear spatial pattern, V3 follows the general patterns and magnitudes of Fan and de Graaf's simulations with highly locally variable WTD. Across the USA's Appalachia, V3's WTD is deep ($>50m$) along steep, high-elevation landscapes, while the rest of the eastern coast has distinctly shallower WTDs at 0-10m. Another major distinction occurs in the large portions of Canada's North surrounding Hudson Bay. In V1, this region has WTDs ranging from 1 to 20 meters, similar to de Graaf's physical-based simulation, whereas V2 predicts almost the entire region to have around zero WTD. While V3 predicts a large portion of this region with nearly zero WTD, some local areas have medium to high WTDs. V3 in this region follows Fan's simulation in terms of the spatial pattern of WTD. Finally, all three machine learning models, in line with Fan and de Graaf's simulations, predict that the southwestern deserts of the USA have extremely deep water table depths ($>70$ m).

The overall spatial patterns of de Graaf and Fan's simulations are similar, although Fan's simulation has much greater local-scale spatial variability. This weakens the spatial correlation between de Graaf and Fan's simulations (0.33). In terms of the magnitude of WTD at the regional scale, Fan's simulation, in line with V3, shows that the lower Mississippi River basin, the coastal region from southern Florida to New Jersey, the midwest United States, and northeastern Canadian wetlands have generally shallow WTDs ($<5m$). At the same time, local areas of deep WTDs also exist along mountainous ridges of the Rockies and Appalachia for both Fan and V3 simulations. Generally, Fan and V3 exhibit high local variability, which arose from simulating the phreatic surface as an extremely damped version of local topography, particularly in mountainous regions (i.e., British Columbia). This pattern matches with most conceptual diagrams of the water table along a hillslope cross section (i.e., see Figure 1 of \citeA{moreno2016modeling}, Figure 1 of \citeA{zang2018numerical}, and Figure 1 of \citeA{rahman2023simulating}). The equilibrium seeking physical equations of Fan and the addition of HAND observations in V3 leads to these models predicting shallow WTD$<5$m in the valleys and deep WTD$>100$m on the mountain ridges. Indeed, WTD strongly follows local-scale elevational differences in both Fan and V3 simulations.

\subsubsection{Comparison among five simulations}

\begin{figure}[h!]
     \centering
     \includegraphics[width=0.55\textwidth]{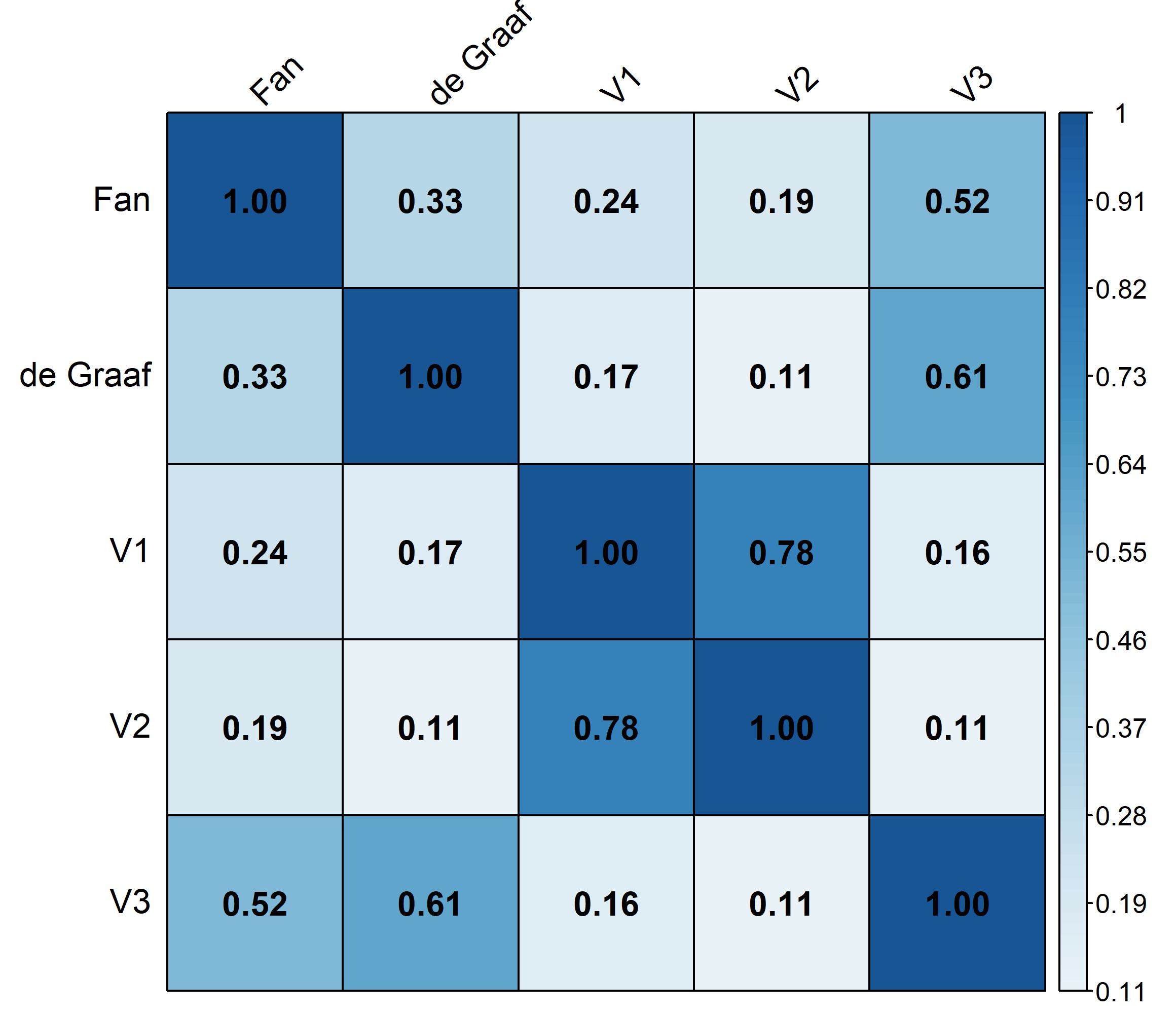}
     \caption{Pixel-by-pixel spatial correlations among the physically-based and machine learning WTD simulations.}
\label{fig:CorPlot}
\end{figure}

Analysis of pixel-by-pixel spatial correlation between each pair of WTD simulations, using all pixels of the study region, depicts there is generally a weak spatial correlation among the two physically-based simulations of WTD ($r \approx 0.33$) across the USA and Canada (Figure \ref{fig:CorPlot}). There is also a weak spatial correlation between the two physically-based simulations of WTD and the V1 and V2 machine learning-based simulations ($0.11<r<0.24$). Interestingly, our V3 simulation is strongly correlated with both physically-based simulations ($0.52<r<0.61$). V1 and V2 are highly correlated with each other ($r = 0.78$), but are less correlated with V3 ($r = 0.16, 0.11$). Such correlations are supported by the continental-scale spatial distribution of simulated WTDs shown in Figure \ref{fig:SimulatedWTD} and the corresponding comparisons made in the previous subsections.

\begin{figure}[h!]
     \centering
     \includegraphics[width=1\textwidth]{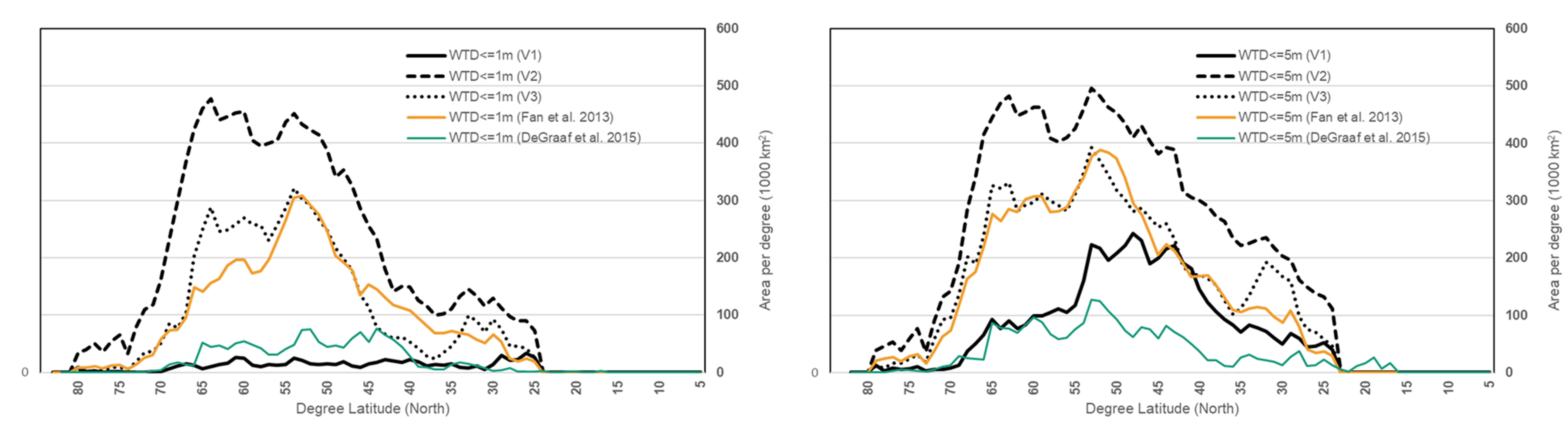}
     \caption{Comparison of  the distribution of areas with water table depths (WTD) less than 1 m (left) and less than 5 m (right) across three machine learning (ML)  and two physically-based WTD simulations. The x-axis represents latitude in 1 degree increments, while the y-axis shows the total area of pixels with shallow WTDs within each 1 degree latitude band.}
\label{fig:Latitude_compare}
\end{figure}
Shallow Groundwater Areas (SGAs) in North America exhibit uneven distribution across latitudes, driven by complex interactions among climatic, hydrologic, and geologic factors (Figure \ref{fig:Latitude_compare}). A strong concordance is observed between the results of V3 and Fan, particularly in sub-polar latitudes. Among the simulations, V1 and DeGraaf display the smallest SGAs, while V2 shows the largest extent.

\subsection{Interpretation of Machine Learning Models}

\begin{figure}[h!]
     \centering
     \includegraphics[width=0.86\textwidth]{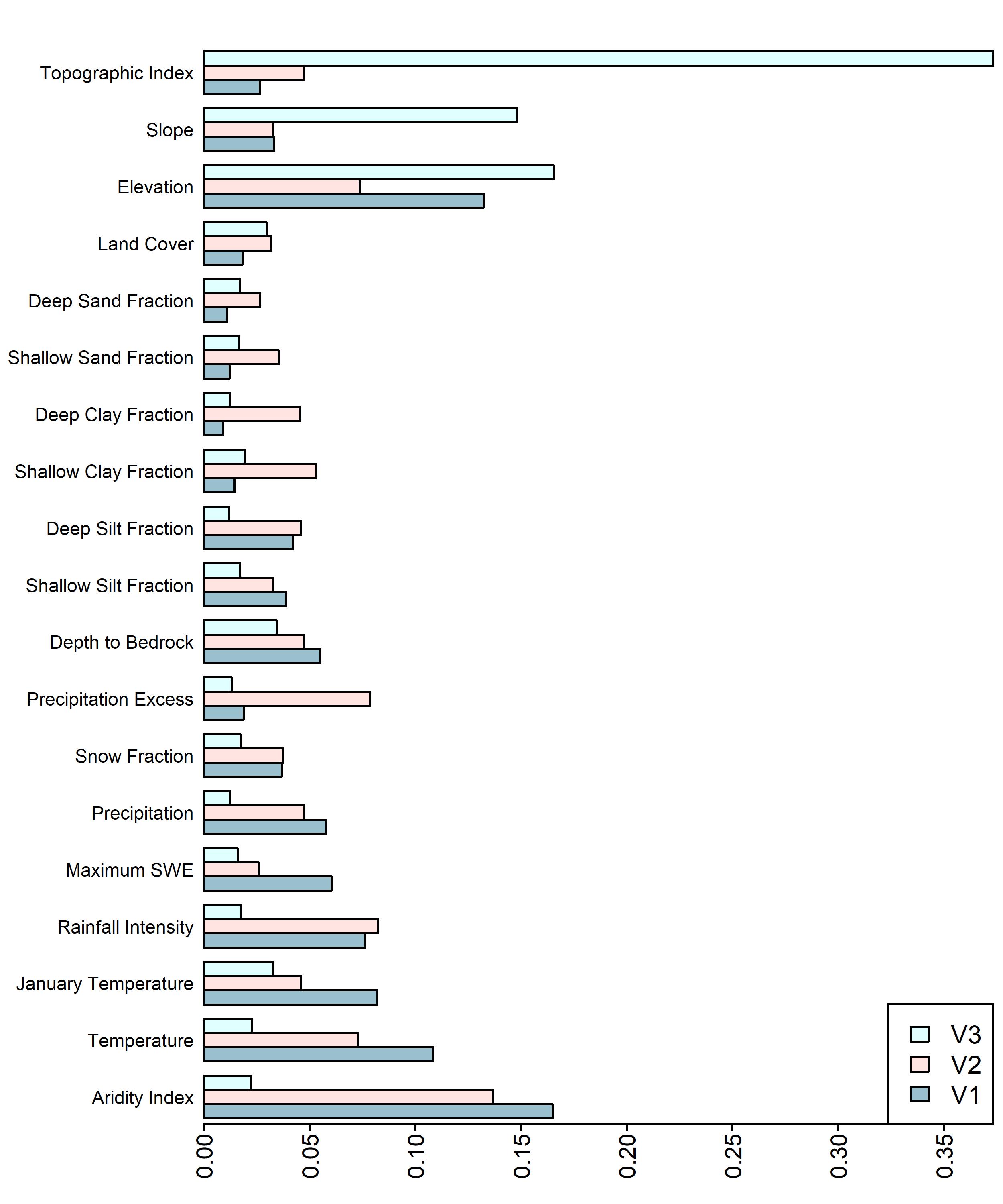}
     \caption{Relative importance of each input variable within each machine learning model (see Section \ref{sec:interpMethods} for the details on the derivation of importance score). The importance scores across a single model add to one.}
\label{fig:ImpPlot}
\end{figure}

This section interprets the importance of the input variables within each of our three machine learning models (Figure \ref{fig:ImpPlot}). V1 strongly relies on climatic features such as temperature and aridity index to explain the variability of real observations of WTD, with both being responsible for over 10\% of the model performance. In total, climatic features have a part in over 60\% of V1's performance, and topographic features explain more than 25\%, with close to 15\% of the impact coming from elevation. V1's simulation and the derived variable importance are highly consistent with \citeA{ma2024water} machine learning simulation of WTD across a large portion of the USA. Land cover and soil properties make up the rest of the model. While land cover and sand (and clay) related fractions contribute the least ($<5\%$), shallow and deep silt fractions and depth to bedrock play moderately strong roles ($5-10\%$). In V2, the aridity index greatly impacts model performance ($>10\%$), while almost all other climatic, topographic and soil-related properties have uniformly moderate ($5-10\%$) effects on V2 model performance.

V3 is almost exclusively reliant on topographic variables ($\approx 70\%$ in total), such as topographic index ($\approx 37\%$), slope ($\approx 15\%$), and elevation ($\approx 17\%$). In total, climatic features explain only 15\% of the model, with January temperature having the largest impact ($\approx 5\%$), while V3 has little (less than $5\%$) reliance on snow fraction, precipitation, precipitation excess, maximum snow water equivalent on the ground, and rainfall intensity. V3 only slightly relies on sand/silt/clay fractions (less than $5\%$), while it is moderately reliant on depth to bedrock ($\approx 5\%$) (Figure \ref{fig:ImpPlot}). 

Overall, V1 and V2 rely on topographic variables (topographic index, slope, and elevation) at about the same level, while V3 is over twice as dependent on these variables. V1 (V3) is strongly (weakly) reliant on climatic information, while V2 gives almost all input variables a more or less uniform score (Figure \ref{fig:ImpPlot}). Land cover does not seem to be a useful variable across all models. Soil-related features such as sand and clay fractions are only given relatively high importance by V2. The stark differences in simulating WTD in mountainous areas between V1 (or V2) and V3, as explained in Section \ref{sec:general_wtd_results}, likely have to do with V3 being strongly driven by local topography instead of climate, while V1 (or V2) is strongly driven by climate. Note that the variable importance metric used in this study is only meant to interpret each model, not the underlying physical processes. Thus, from this analysis, we can only conclude how heavily each model uses each variable to explain the variability of target observations, not how strongly each variable is related to WTD in general (see \citeA{janssen2023ultra} for details). This important dichotomy is illustrated by the fact that each model gives vastly different importance scores and the ground-truth importance scores are currently unknown. Further, it is important to note that these importance scores are only related to the spatial heterogeneity of WTD, and it does not consider temporal variation.

\subsection{Relative Comparisons of Models Predictive Performances}
\label{sec:result_predperf}

This section compares each model's ability to predict WTD at pixels wherein unseen real and proxy observations are located within each ecoregion and throughout the study region. Mean absolute error (MAE) and Pearson correlation were used as performance metrics to evaluate model performance in predicting observed or expected WTDs at unseen locations. A summary of predictive performance results for each ecoregion and the entire study region can be found in Table \ref{tab:pred_perf_res}. 

Along the Temperate Sierras (13), MAE-WB values are highest, meaning all models perform relatively poorly (MAE-WB $>2$) in predicting permanent water body locations. This ecoregion exhibits the largest elevation across the study region (median of 2019m). All models, except V2, perform poorly in predicting permanent water bodies along the North American Desert (10), which has the highest aridity among all ecoregions and the second-highest elevation. Again, all models, except V2, perform poorly in predicting permanent water bodies in Mediterranean California (11), which has the second highest aridity. In the Eastern Temperate Forest (8), with the second lowest average elevation, V2 and V3 perform exceptionally well (MAE-WB$\approx 0.1$), and Fan (MAE-WB$\approx 0.3$) performs very well, in predicting small water bodies. Similarly, in the Great Plains (9) and Northern Forest (5), V2 and V3 perform exceptionally well and Fan performs very well. Along the Northwestern Forested Mountains (6) and Marine West Coast Forest (7), V2 is the only simulation with a strong performance in predicting permanent water bodies (MAE-WB$\approx 0.1$). Unsurprisingly, the region with the lowest average observed WTD (i.e. the tropical humid forests of Florida (15)) also has the lowest error in locating permanent water bodies across all models, with V2 and V3 performing exceptionally well (MAE-WB $\approx 0$). Throughout the entire study region, our V2 (MAE-WB=0.1) simulation best predicts the unseen interior locations of permanently wet surface water bodies. V3 (MAE-WB=2.42) and Fan (MAE-WB=2.54) simulations also perform relatively well, while V1 (MAE-WB=5.38) and de Graaf (MAE-WB=15.36) simulations are relatively poor in this respect.  

\begin{table}[tp!]
 \caption{The performance of each model on "unseen" real observations of water table depth (OBS) and unseen interior pixels of permanently wet water bodies (WB) across the entire study area (All) and within each ecoregion. The performance is evaluated using mean absolute error of simulated WTD at interior locations of permanently wet surface water bodies wherein zero WTD is expected (MAE-WB), mean absolute error of simulated versus observed WTD (MAE-OBS), and the correlation between simulated and observed WTD (Corr-OBS). Ecoregions include: Taiga (3), Northern Forests (5), Northwestern Forested Mtns (6), Marine West Coast Forests (7), Eastern Temperate Forests (8), Great Plains (9), North American Deserts (10), Mediterranean California (11), Temperate Sierras (13), Tropical Humid Forests (15). The best result of each row is bolded.}
  \centering
  \begin{tabular}{lllllllll}             \\
     \hline
    \textbf{Region} & \textbf{Real Obs} & \textbf{Eval} & \textbf{Fan} & \textbf{de Graaf} & \textbf{V1} & \textbf{V2} & \textbf{V3}\\
     \hline
    (3) & 483 & MAE-WB  & 2.53 & 49.23 & 5.58 & \textbf{0.04} & 2.83 \\
    & & MAE-OBS & \textbf{3.93} & 29.19 & 7.38 & 6.69 & 28.71 \\
    & & Corr-OBS & 0.18 & 0.31 & \textbf{0.50} & 0.43 & 0.05\\
    \hline
    (5) & 48140 & MAE-WB & 0.91 & 17.07 & 3.90 & \textbf{0.06} & 0.32 \\
    & & MAE-OBS & 8.54 & 30.62 & \textbf{3.96} & 4.64 & 6.10 \\
    & & Corr-OBS & 0.03 & 0.09 & \textbf{0.54} & 0.51 & 0.29 \\
    \hline
    (6) & 27406 & MAE-WB & 4.56 & 91.04 & 6.62 & \textbf{0.09} & 5.52 \\
    & & MAE-OBS & 21.87 & 152.07 & \textbf{11.83} & 12.20 & 16.26 \\
    & & Corr-OBS & 0.18 & -0.07 & \textbf{0.73} & 0.71 & 0.47 \\
    \hline
    (7) & 21301 & MAE-WB & 9.90 & 194.36 & 5.09 & \textbf{0.10} & 11.97 \\
    & & MAE-OBS & 11.97 & 65.69 & \textbf{5.02} & \textbf{5.02} & 6.95 \\
    & & Corr-OBS & 0.13 & 0.05 & \textbf{0.60} & \textbf{0.60} & 0.38 \\
    \hline
    (8) & 122212 & MAE-WB & 0.30 & 3.66 & 3.04 & \textbf{0.10} & \textbf{0.10} \\
    & & MAE-OBS & 7.16 & 23.02 & \textbf{2.73} & 3.25 & 4.22 \\
    & & Corr-OBS & 0.18 & 0.09 & \textbf{0.62} & 0.52 & 0.28 \\
    \hline
    (9) & 228570 & MAE-WB & 0.79 & 15.24 & 4.96 & \textbf{0.12} & 0.38 \\
    & & MAE-OBS & 13.33 & 39.20 & 8.62 & \textbf{8.47} & 9.22 \\
    & & Corr-OBS & 0.38 & 0.20 & \textbf{0.70} & 0.69 & 0.64 \\
    \hline
    (10) & 76231 & MAE-WB & 8.12 & 139.18 & 9.30 & \textbf{0.53} & 4.67 \\
    & & MAE-OBS & 23.36 & 94.55 & \textbf{14.28} & 15.32 & 17.25 \\
    & & Corr-OBS & 0.37 & 0.14 & \textbf{0.73} & 0.69 & 0.55 \\
    \hline
    (11) & 16278 & MAE-WB & 7.76 & 116.09 & 9.55 & \textbf{0.65} & 4.57 \\
    & & MAE-OBS & 14.74 & 72.41 & 9.10 & \textbf{8.68} & 12.14 \\
    & & Corr-OBS & 0.24 & 0.08 & \textbf{0.72} & 0.71 & 0.49 \\
    \hline
    (13) & 2858 & MAE-WB & 9.96 & 202.93 & 22.33 & \textbf{2.19} & 9.70 \\
    & & MAE-OBS & 39.20 & 479.42 & \textbf{29.21} & \textbf{29.21} & 29.76 \\
    & & Corr-OBS & 0.10 & -0.02 & \textbf{0.57} & \textbf{0.57} & 0.19 \\
    \hline
    (15) & 857 & MAE-WB & 0.29 & 1.93 & 1.67 & 0.05 & \textbf{0.03} \\
    & & MAE-OBS & 1.46 & 1.36 & \textbf{0.96} & 1.31 & 1.31 \\
    & & Corr-OBS & 0.09 & 0.10 & 0.12 & \textbf{0.26} & 0.09 \\
    \hline
    All & 541418 & MAE-WB & 2.54 & 40.78 & 5.38 & \textbf{0.10} & 2.42 \\
    & & MAE-OBS & 13.47 & 52.90 & \textbf{7.81} & 8.18 & 9.61 \\
    & & Corr-OBS & 0.40 & 0.21 & \textbf{0.75} & 0.71 & 0.60 \\
     \hline
  \end{tabular}
  \label{tab:pred_perf_res}
\end{table}

Table \ref{tab:pred_perf_res} clearly indicates that the three machine learning-based simulations provide stronger corelation between simulated and observed WTD (Corr-OBS between 0.6-0.75) compared to the two physically-based simulations (Corr-OBS between 0.21-0.4). Within ecoregions, both MAE and correlation metrics depict that the three machine learning-based simulations perform better in predicting unseen real observations of WTD, except in Taiga (3) with a small number of observations, where the Fan simulation performs best. Although they are never better than all ML models, except in the case of Taiga above, the simulations of \citeA{de2015high} and \citeA{fan2013global} sometimes perform better than one set-up of machine learning model. For example, in terms of MAE of water body locations, Fan simulations outperform V1 in almost all regions and Fan outperforms V3 for this metric in Taiga, Marine West Coast Forests, and Eastern Temperate Forests. Further, in the sparsely observed region of Taiga, both de Graaf and Fan outperform V3 in terms of correlation with observations. In predicting WTD at unseen wells across the entire study region, V1 performs the best among all models with a correlation of 0.75 and MAE of 7.81. This is not surprising since it is the only simulation exclusively driven by real observations of WTD. Although most of V2's training data comes from proxy observations (and not real observations of WTD), its performance when predicting unseen real observations degrades very little, with a correlation of 0.71 and an MAE of 8.18. The errors and accuracies among these two models and within each ecoregion also correspond closely. V3, generally, performs worst among the three machine learning models in predicting real observation of water table depth while still performing better than the two physically-based models. 

\begin{figure}[h!]
\captionsetup[subfigure]{aboveskip=-1pt}
     \centering
     \begin{subfigure}{0.491\textwidth}
         \centering
         \includegraphics[width=\textwidth]{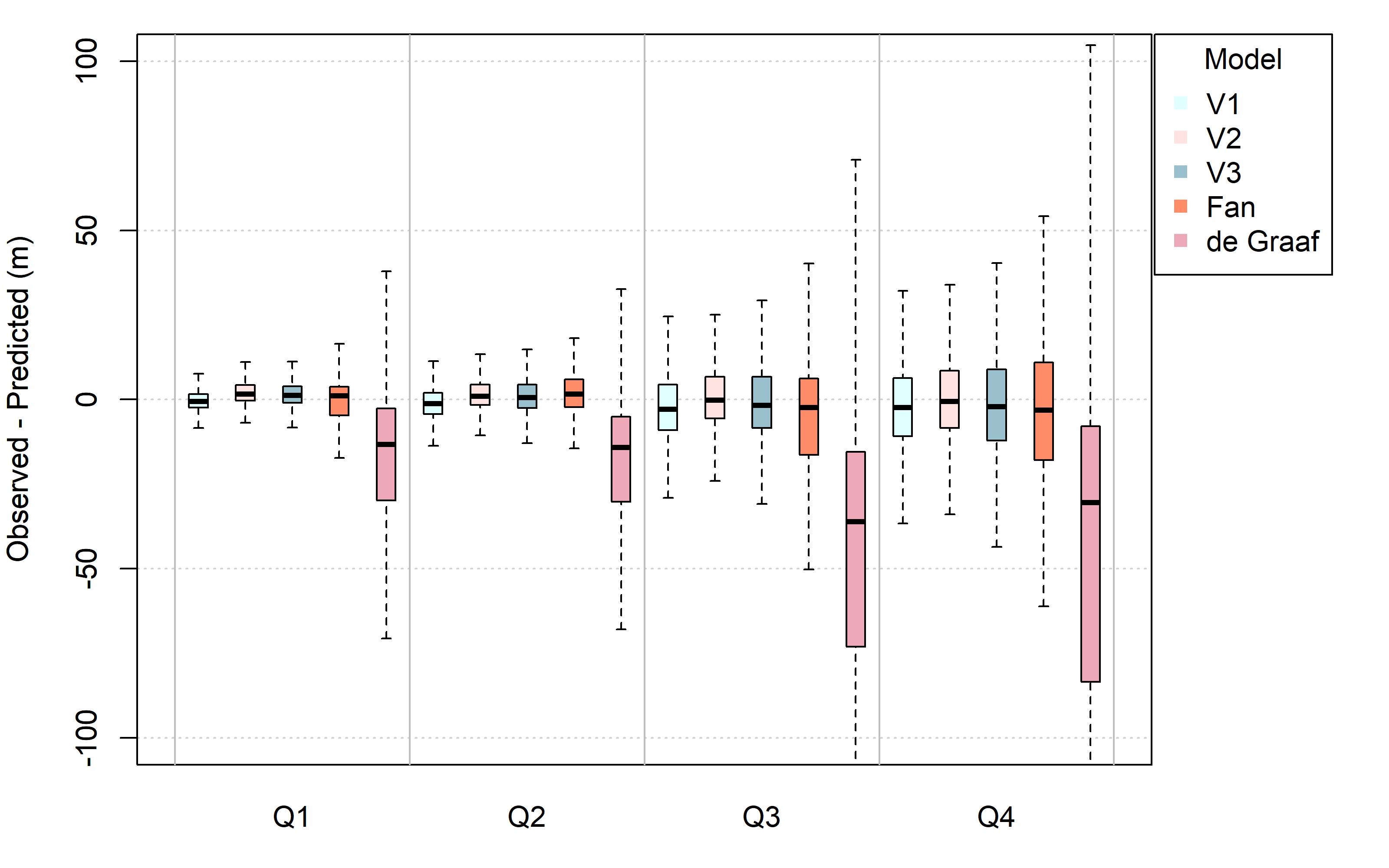}
         \caption{Aridity Index}
         \label{fig:eval_aridity}
         \setcounter{subfigure}{1}
         \includegraphics[width=\textwidth]{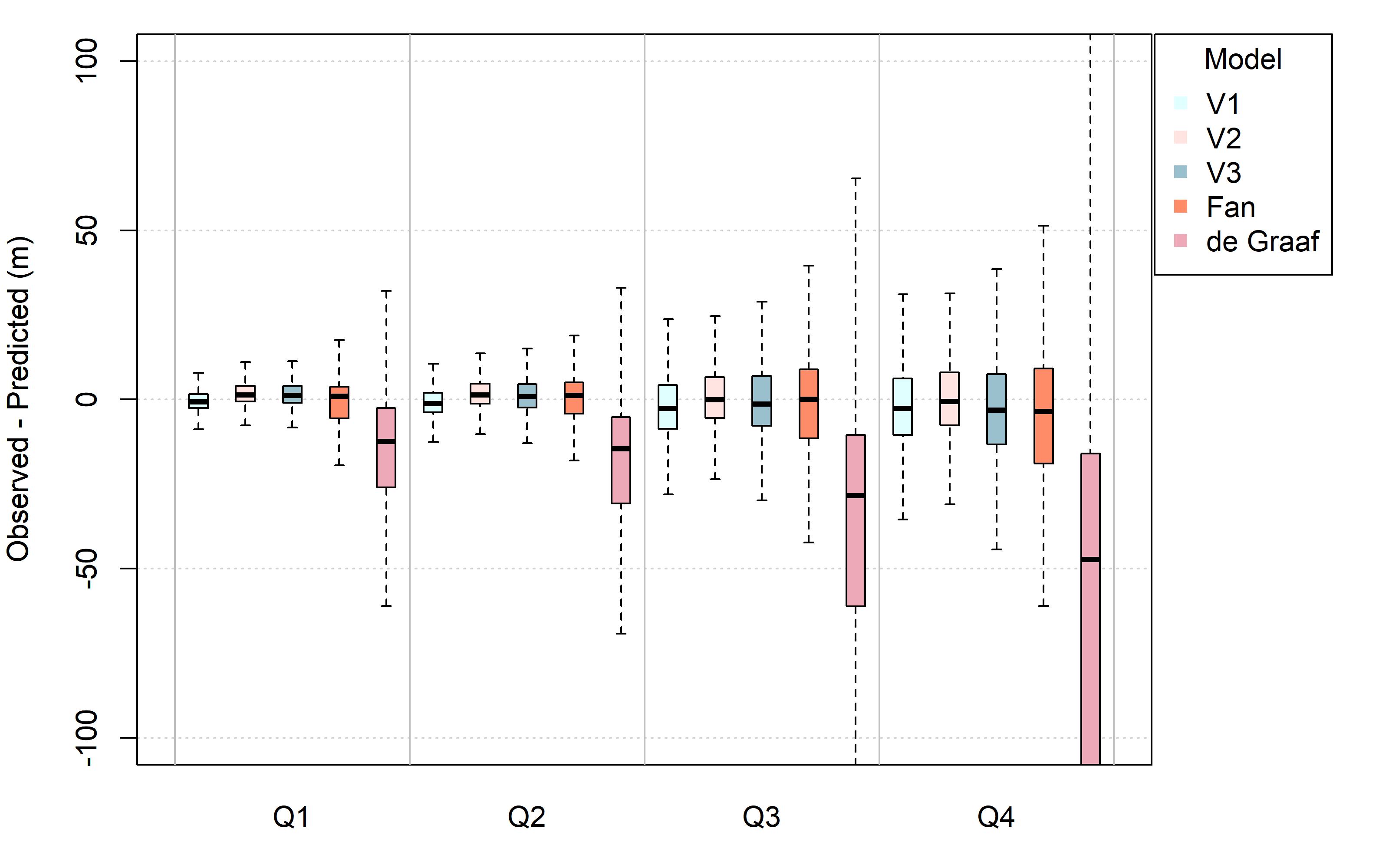}
         \caption{Elevation}
         \label{fig:eval_elevation}
         \setcounter{subfigure}{2}
         \includegraphics[width=\textwidth]{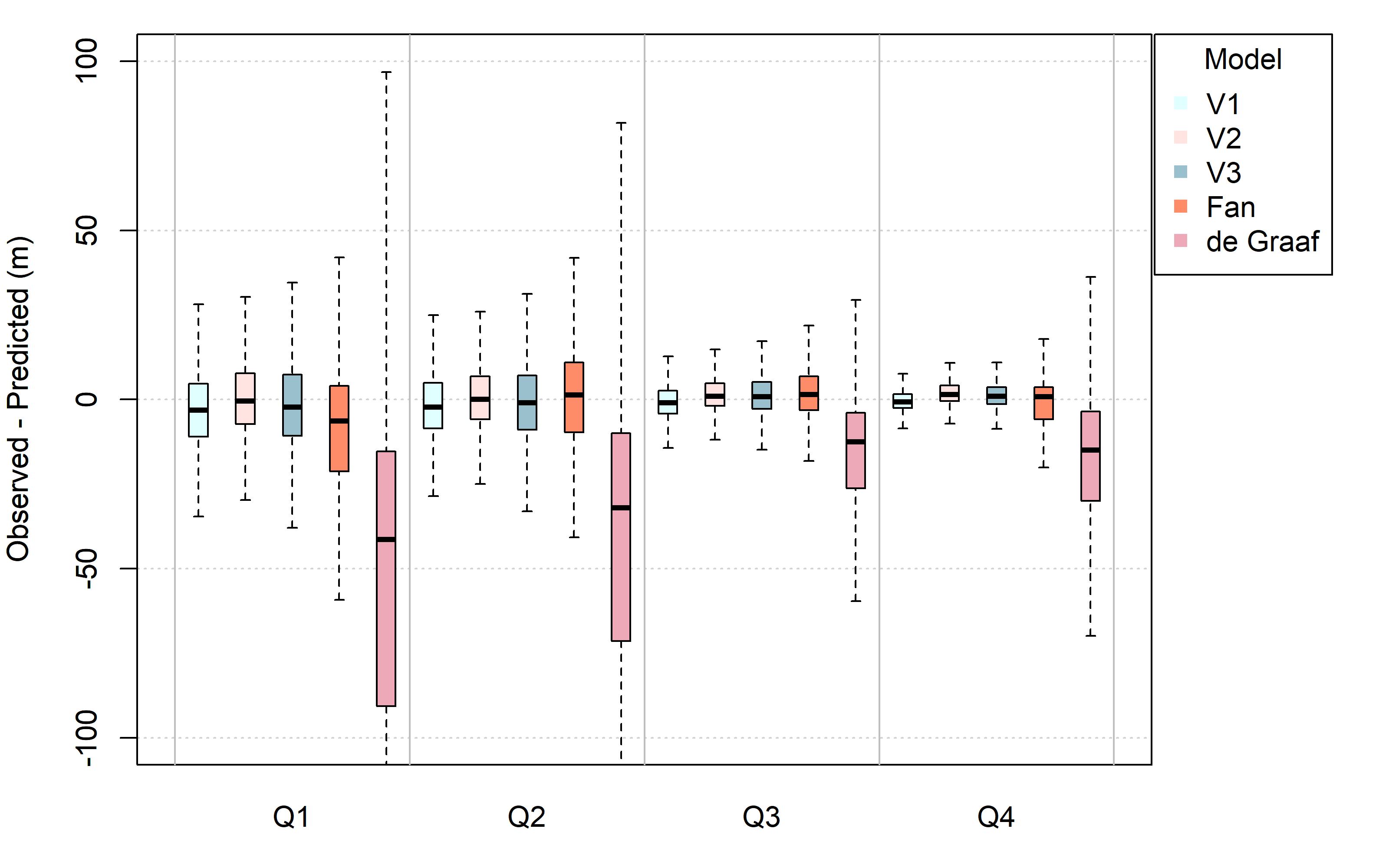}
         \caption{Precipitation}
         \label{fig:eval_precip}
         
         \end{subfigure}
     \hfill
     \begin{subfigure}{0.491\textwidth}
         \centering
         \includegraphics[width=\textwidth]{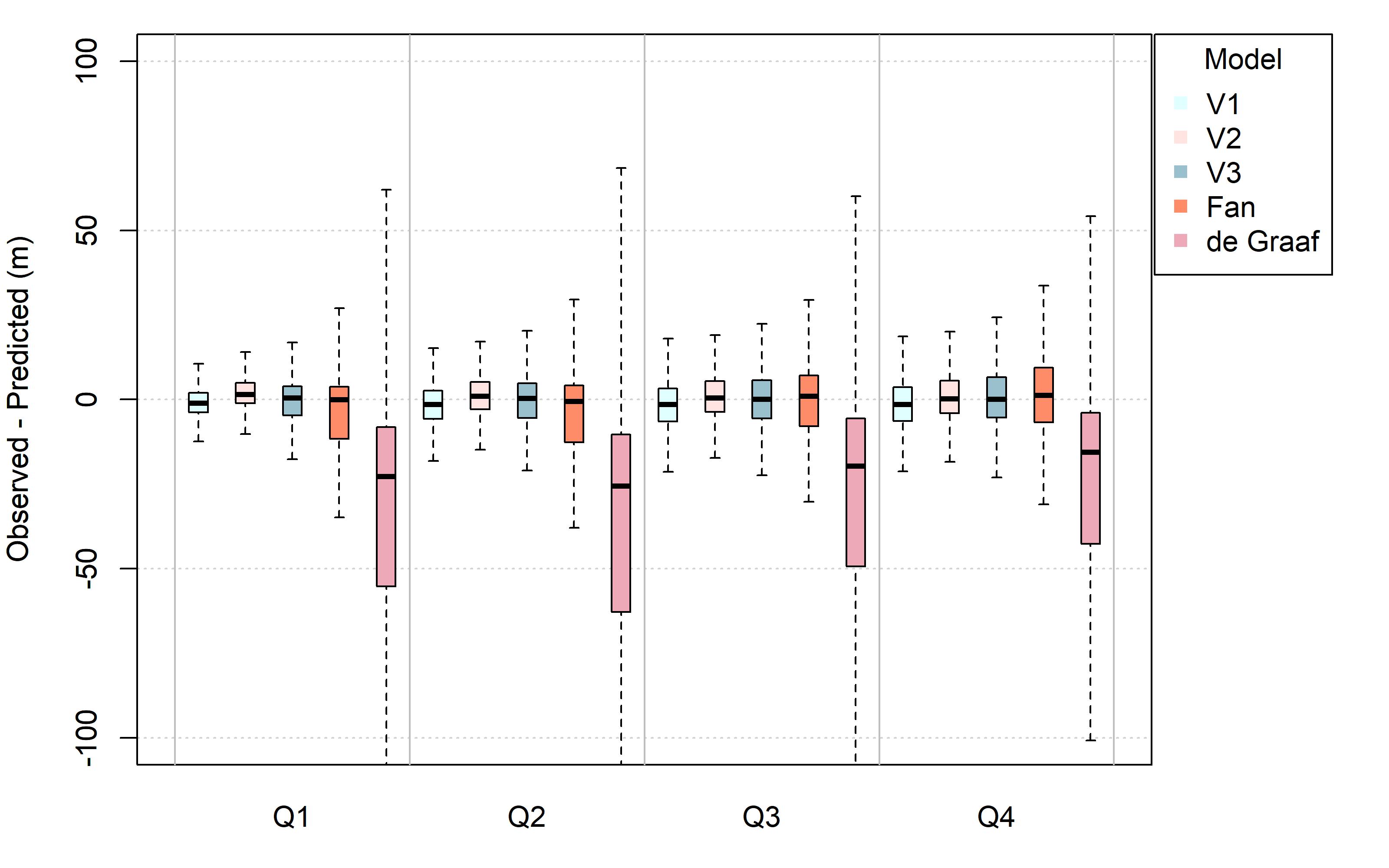}
         \caption{Depth to Bedrock}
         \setcounter{subfigure}{4}
         \label{fig:eval_DTB}
         \includegraphics[width=\textwidth]{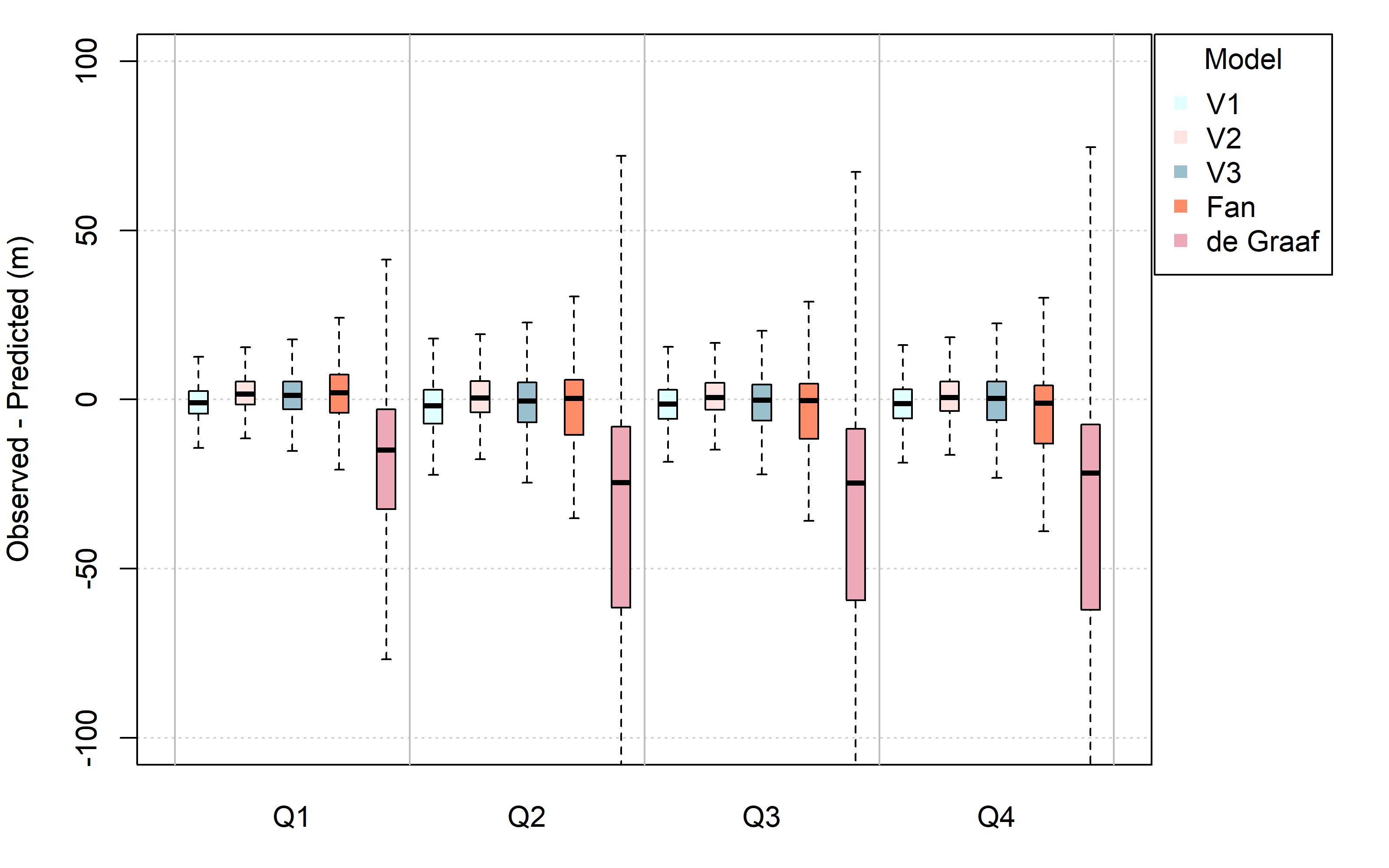}
         \caption{Shallow Sand Fraction}
         \label{fig:eval_ShallowSand}
         \setcounter{subfigure}{5}
         \includegraphics[width=\textwidth]{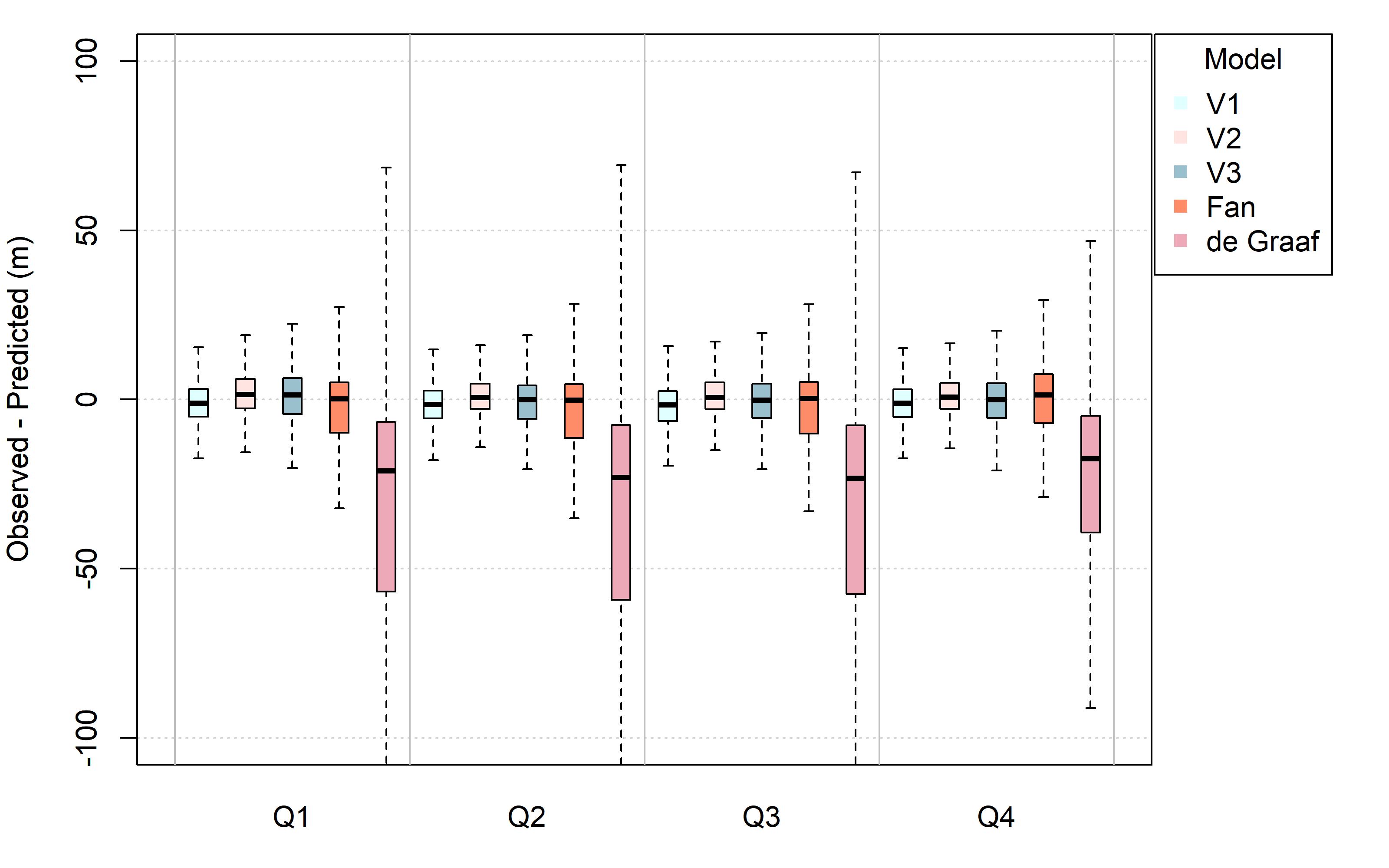}
         \caption{Deep Silt Fraction}
         \label{fig:DeepSilt}
         \setcounter{subfigure}{6}
          \end{subfigure}
     
        \caption{Distribution of residuals (Observed - Predicted WTD) across four quartiles of selected globally available inputs and across the five machine leaning and physically-based models.}
\label{fig:eval_byvar}
\end{figure}

The residuals (i.e. Observed - Predicted WTD) from \citeA{de2015high} are substantially larger in magnitude and also more heavily biased (predictions are too deep) compared to those of Fan, V1, V2, and V3  (Figure \ref{fig:eval_byvar}). Across most quartiles of input attributes, there are very few differences in the residuals between V1 and V2. V3 generally has slightly higher error compared to V1 and V2, while Fan generally has higher error compared to V3.

In terms of specific predictor variables, Figure \ref{fig:eval_byvar} shows that aridity index, elevation, and precipitation are strong predictors of residual magnitudes. In particular, WTD errors tend to be larger as aridity or elevation increase. Across all models, as regions become drier (from the fourth quartile Q4 to the first quartile Q1 of precipitation), the errors between modelled WTD and real well observations tend to increase. Higher depth to bedrock (comparing Q1 with Q4) seems to be associated with higher errors in the machine learning-based models and Fan's simulation. Higher shallow sand fraction tends to lead to slightly higher errors in all models. 

\section{Discussion}
\label{sec:discussion}

This paper presented three physically constrained machine learning (XGBoost) simulations of WTD. The three models (V1, V2, and V3) were built by sequentially adding proxy observations relevant to WTD (see Section \ref{sec:threeModels}). Unseen real and proxy observations of WTD, which were not used for training and validation of machine learning models, were then employed to explore and compare the performances of the three machine learning simulations and  two physically-based available simulations of WTD across ten ecoregions of North America. Overall, all five models poorly predict WTD across three ecoregions with the highest aridity and elevation (Temperate Sierras, North American Desert, Mediterranean California). In the rest of the USA and Canada, along with the locations with available proxy data on surface water bodies V2, V3, and (in some eco-regions) Fan's simulation perform relatively strongly. Where real observations of WTD exist, V1, V2, and V3 perform relatively stronger than physically-based models in following the variability or magnitudes of well-based observations. The interpretation of the three machine learning models suggested that the V3 model strongly used the topographic wetness index (TWI) information to simulate WTD, while V1 strongly relies on climatic information. In this regard, V1 resembles \citeA{ma2024water}'s machine learning simulations of US WTD, which used random forests trained on only well-based observations of WTD. 

One may argue that the information that V3 learnt from the data and used to predict WTD corroborates well with classical physically-based models (e.g., TOPMODEL \cite{beven2021history}), in which TWI dominantly controls the spatial pattern of WTD. Recent hillslope-scale experimental works in both wet and arid mountainous landscapes also suggested that TWI dominantly controls the location of WTD from the hillslope ridge line (where TWI is small and WTD is deep) to the hillslope toe (where TWI is relatively large and WTD is shallow) (e.g., \citeA{van2024ir,karlstrom2023state}). Among the three machine learning simulations, V3 not only somewhat strongly follows WTD observations and predicts interior wet pixels of surface water bodies, in most of North America's ecoregions, it also generates predictions which are highly spatially correlated with simulations from Fan's and de Graaf's physically-based models, particularly along mountainous regions. Indeed, V3 provides a good balance between what (real and proxy) data dictates and what physically-based models suggest. However, both data and model structures, are uncertain and difficult to verify. These uncertainties will be critically discussed below, guiding future works on data collection and compilation as well as model development and verification in groundwater hydrology. 

\subsection{Uncertainty in The Observations of WTD}
\label{sec:uncertain_data}

To assess the reliability of different WTD simulations, we must rely on historical observations of well-based water levels. However, these real observations may have associated errors and, under some circumstances, may not represent the water table depth meaningfully \cite{taucare2024alarming}. This is particularly important as the model evaluation statistics can depend strongly on the sample of observations on which one chooses to evaluate the model (see \citeA{reinecke2020importance} and Table \ref{tab:pred_perf_res}). In the remainder of this subsection, we explore uncertainties related to the real observations of WTD and how they can affect the simulations of WTD and the verifiability of those simulations.  

Observations of WTD are susceptible to systematic errors due to vertical hydraulic gradients \cite{jasechko2020california}. It is fairly well known that water tapped within confining layers can experience additional pressure heads, leading to extremely shallow measurements of well water levels. In these systems, the vertical gradient would be upwards, and the true water table would be far deeper than what the readings from well observations convey \cite{vincent2013hydraulic,hilton2023widespread}. Very few well observations have indicators (and labels) of whether or not a confined or unconfined aquifer is tapped, further adding to the uncertainty of observations of WTD. The upward vertical gradient issue can even arise in unconfined aquifers, especially in the valley bottoms of steep terrains \cite{hilton2023widespread}. Furthermore, downward hydraulic gradients could be common in some landscapes \cite{elcci2003detrimental,jasechko2020california,gabrielli2018contrasting}, where the WTD could be measured far below the true water table. This can happen due to fairly independent causes such as long well screens \cite{elcci2003detrimental}. Theoretical and numerical simulations conducted by \citeA{elcci2003detrimental}, as well as an experimental study conducted  in California by \citeA{jasechko2020california} depicted this phenomenon. Indeed, well levels are only a good representation of the water table when the well screen is placed at (or close to) the true water table or where perfectly hydrostatic conditions exist \cite{vincent2013hydraulic,jasechko2020california}. 

Human error could also cause some types of data uncertainty. Across our study region, several real WTD observations are unreliable. Over 100 USGS observations showing a water level deeper than the deepest well ever drilled in North America and thousands of USGS observations showing negative WTDs. Various other obvious errors have been detected in other works as well \cite{jasechko2020california}. While these detected observations are wrong and should be removed before training or validating a model (as was done in our paper), other observations may be equally wrong, but impossible to detect and remove. In addition to large human errors, which may be present in a minority of observations, random observational errors can occur everywhere \cite{vincent2013hydraulic}. For example, \citeA{silliman2000effect} found that noise in water level measurements is high enough to prevent the estimation of the vertical hydraulic gradient.

Pumping has greatly affected groundwater levels across the US, especially in the last hundred years \cite{hilton2023widespread,luczaj2017aquifer}. Groundwater pumping can have an outsized impact on lowering the water table in warm, low-elevation landscapes due to the higher likelihood of agricultural uses \cite{fan2013global,gleeson2021gmd}. Pumping can affect WTD observations near and in the surrounding areas where pumping occurs. The observations collected near pumping areas do not necessarily reflect the natural and permanent WTD \cite{jasechko2020california}. With data from California, \citeA{jasechko2020california} found that pumping can cause drawdowns of 10s of meters, and groundwater level recovery after overexploitation often takes decades or centuries in arid regions and may even be permanent \cite{van2009recovery,sommer2011resilience}. Even after the use of a well has ceased, an artesian well may continue to extract groundwater and affect observations of WTD \cite{hilton2023widespread}.

Temporal variability of the water table may hinder our ability to identify static WTD at a particular location \cite{molenat2005model}. A large portion of real observations in the USA and Canada ($\sim 81 \%$) were taken at one time, with no follow-up observations at that location. Water levels can fluctuate due to long-term or oscillatory changes in climate, tides, runoff, barometric pressure, mass loading, or pumping \cite{condon2021global,rust2019understanding,rasmussen1997identifying,flickinger2020water,molenat2005model,hayashi2016hydrology,jasechko2024rapid}. Appraisal of USGS data depict, for example, in a well near Seco Creek, Texas, the WTD fluctuated by over 50 meters in less than two decades of observations. For a USGS groundwater well in San Bernardino, California, yearly oscillations occur with an amplitude of nearly 10 meters, with an additional decreasing trend from about 30 meters in 2005 to 60 meters in 2020. Such temporal variation of WTD could also happen in regions with typically shallow WTD, such as the prairie pothole region of central USA and Canada. In this region, the water table may fluctuate by over 6 meters in less than a decade, even without pumping influences, as exemplified by \citeA{hayashi2016hydrology}. In aquifers worldwide, long-term trends of 10 m to over 100m in observed WTD can be caused by excessive pumping \cite{jasechko2024rapid,lopes2006water}. Even when we exclude pumping effects, natural seasonal variations were shown in Nevada to vary WTD by about 6m \cite{lopes2006water}. 

Finally, and perhaps most importantly, large-scale sampling biases across the USA and Canada have affected the overall spatial pattern of available observations of WTD \cite{reinecke2023global}. Hydrologists and engineers typically dig wells where there is likely water (i.e., river valleys and productive aquifers). If a well is dug to some maximum acceptable depth and no water is found, no WTD measurement can be taken. Hence, deep WTD observations are not widely available across North America.  Furthermore, the WTD observation dataset can be biased towards observations in low-elevation areas with large populations and moderate climates instead of dry (or wet) high-elevation areas \cite{boerman2022comparing,lopes2006water}. Therefore, the overall distribution of observations of WTD, available for modeling studies, may be shallower than the actual distribution of WTD across the USA and Canada \cite{de2015high}. 

Note that  the need for high quality data is not limited to development of machine learning models, as the mentioned data uncertainties can equally affect the verifiability of physically-based and machine learning models. We further discuss the future needs and directions to access high quality observational or proxy data of WTD in Section \ref{sec:future}.

\subsection{Uncertainty in Machine Learning-based simulations of WTD}
\label{des:uML}

The rapid adoption of machine learning within hydrology has led to several important advancements due to these algorithms' ability to easily model and predict hydrology's complex and non-linear processes. Yet, several shortcomings still remain for groundwater modelling using machine learning frameworks, as evidenced in our paper by the lack of strong fit among models as well as the lack of strong fit between simulations and observations of WTD. The three primary shortcomings are the overreliance on observational data, equifinality of model structure, and the inability to directly model physical processes \cite{ISTALKAR2023Value, beven2002towards}. While all these shortcoming are also partially relevant to large-scale physically-based simulations of WTD, here we mainly focus on their effects on machine learning model development in groundwater hydrology.   

The three machine learning-based groundwater models analyzed in our work are based on the same powerful machine learning algorithm, XGBoost, and the same set of physically-based constraints. Yet each model provides locally distinct simulations of WTD (Figure \ref{fig:SimulatedWTD}). Furthermore, the model performance against unseen real observations of WTD is not as strong as the performance obtained in other areas of hydrology (e.g., streamflow prediction). Using similar tree-based machine learning methods, \citeA{koch2019modelling} and \citeA{koch2021high} found that the highest $R^2$ against unseen observations they could achieve across Denmark was about 0.55. Further, \citeA{ma2024water} showed that random forests, could obtain 0.65 test set NSE with respect to log-transformed WTD along a large portion of the USA . Our test set $R^2$ (NSE) against unseen real observations of WTD across the entire USA and Canada was quite similar for V1 (0.56) and V2 (0.52). 

Typically, stronger performance against observational data is preferred, and systematic biases and large random errors in predictions suggests that the developed model misrepresented the groundwater system. However, due to the considerable uncertainties of the observational data (see Section \ref{sec:uncertain_data}), it is unclear if the models that make the best predictions against real observations of WTD (V1 and V2) are truly the best. Experimental and modeling studies revealed WTD is usually extremely deep across steep (and wet) portions of the Rocky Mountains of Canada, in line with the simulations made by V3 \cite{chen2020towards,fan2013global,de2015high,smerdon2009approach}. Yet in the corresponding ecoregions (6 and 7), the performance of V3 is worse than V1 and V2 (but still better than the physically-based simulations), if one considers the real observations of WTD as ground truth (Table \ref{tab:pred_perf_res}). Without well-regulated observations that reflect the true WTD, we, as groundwater hydrologists, will forever be blind to the ground truth. With many parameterization (or model structure) options available with machine learning algorithms (similar to physically-based models), only reliable observations can help us choose the correct path. If we knew the exact physical equations driving the water table location at a coarse scale (e.g., 500 m), knew accurate subsurface parameters at all locations, and could engineer a model that perfectly incorporates these equations and parameters without simplification, the reliance on  observational data would be less, and one could ignore biased training data. But clearly, these coarse-scale ground-truth equations and parameters have yet to be found. 

Equifinality can have a profound negative effect on the trustworthiness of machine learning simulations. During hyperparameter tuning and training, we noticed that there are likely many XGBoost parameter sets spanning wide ranges of the parameter and hyperparameter space, with satisfactory performance on the validation/test set. Indeed, as other studies showed, the most complex and least interpretable models often have the best predictive capability \cite{nearing2021role,yang2021reliability}. To manage equifinality, one solution is to add physical realism by adding physical constraints to machine learning algorithms, as we did in our paper. However, this practice reduced our models' predictive capability, as noted at the beginning of this section and as shown in other studies as well \cite{yang2021reliability}. One possible explanation is that adding physical realism to part of a machine learning model, in the form of monotonically constraining relationships of some variables, could diminish the physical realism of other more uncertain parts of the model as exemplified in \citeA{yang2021reliability}. To properly constrain machine learning models, a better understanding of all physical (potentially complex and nonlinear) relationships among inputs and output data and/or more diverse observations relevant to different processes connecting the input to output is needed \cite{beven2002towards}. 

Machine learning models can far ``outperform" physically-based models on test set data (as shown in our paper and other hydrology related disciplines , e.g., \citeA{nearing2021role}). However, given the high sampling biases of our real observations of WTD and observational data uncertainties  (see Section \ref{sec:uncertain_data}), across unmeasured or unmeasurable locations, machine learning is still not guaranteed to ``physically outperform" fine-resolution physically-based models. Although we attempted to enhance the physical realism of our machine learning models through the use of additional proxy observations and physics-based constraints, we still expect future works could further enhance the physical realism of machine learning models in groundwater hydrology. We will discuss these points further in the next section.

\subsection{Future Directions}
\label{sec:future}

Currently, our knowledge about water table locations is highly uncertain and stored in hearsay and newspaper reports \cite{jasechko2020california}. This is mainly because potentially uncertain and equifinal machine learning (or even physically-based) models cannot be verified with potentially uncertain (and non-representative) observations of WTD. Here, we discuss future directions for improving the physical realism and verifiability of machine learning models in groundwater hydrology.   

The physical realism of machine learning models could be enhanced by incorporating known physical laws of groundwater hydrology (e.g., Darcy's Law). However, there is no guarantee that including these laws would \textit{verifiably} increase the physical realism of models at the scale of interest. This is also true for the physically-based models where laboratory-scale physical laws may not control pixel-scale groundwater processes.  How climate, topography, and geology interact at local and regional scales could have an outsized effect on the pixel-scale WTD. Indeed, groundwater flow is not only driven by the landscape attributes at a given location but also by that of surrounding areas \cite{grabs2009modeling}. For example, blowing snow, runoff over frozen ground, and lateral groundwater movement have pronounced causal effects on groundwater levels several kilometers away \cite{hayashi2016hydrology}. Yet, such complex and spatially-varying interactions and the functional relationships connecting landscape attributes (locally and regionally) to WTD are understudied and remain unincluded in current machine learning models. Once such interactions and functional relationships are defined, they could be used to select among many candidate models or incorporated a priori in machine learning models.

Reliable verification among competing machine learning models could be made possible by testing how competing groundwater models simulate known causal relationships between the drivers of WTD (e.g., precipitation, topographic index) and WTD (or hydraulic gradient). This can be done using metamorphic testing if we know a priori the function and form of causal relationships among the drivers and WTD \cite{yang2021reliability}. The monotonic physical constraints used in our paper (see Section \ref{sec:montone}) are a simple form of such causal relationships. However, in reality, the functional forms of such relationships at local and regional scales can be more complex than what we used in our paper. Hence, future experimental or statistical inference studies could focus on exploring the functional form of such causal relationships at different scales. Another alternative to verify models could be the use of virtual experimentation wherein the competing machine learning models virtually simulate land use and climate alterations to choose which model more reliably replicates known behaviour \cite{kirchner2006getting}.  

Process-oriented indices, data, and markers with information relevant to WTD can improve machine learning-based groundwater models through different pathways. They can be used as (1) input variables to the model, as done for the topographic index in our paper, (2) proxy observations for model training, as done for the HAND and the probability of surface inundation in our paper, (3) model constraints or boundary conditions, or (4) as an (in)direct verifier to select the model structure that makes good predictions for the right reasons. For example, previous works showed that the incorporation of information on the landscape's hydrologic functioning (i.e. how landscapes store and transmit groundwater), through novel scientific indices, could enhance the generalizability of machine learning-based groundwater-surface water interaction models \cite{janssen2021hydrologic}. Topography-based indices and satellite data, reflecting how landscapes have evolved, can be derived at fine resolution and ultimately can be used to enhance the capabilities of machine learning models. Indices such as height above the nearest drainage or water body (HAND), horizontal distance from the water body, and topographic index turned out to be the most important topography-based drivers of WTD in previous studies \cite{aagren2014evaluating,gabrielli2018contrasting,koch2019modelling,koch2021high,grabs2009modeling}. Distance to hydrogeologically significant features such as alluvial-fan or consolidated rock may also be important \cite{lopes2006water}. One could also go beyond our simple monotonic constraints by more closely coupling physically-based and machine learning models. Perhaps there is a way to incorporate differentiable modelling framework \cite{shen2023differentiable} within groundwater modelling by increasing the flexibility of rigid groundwater flow laws, or by constraining data-driven methods to only allow for information/groundwater to flow according to pressure gradients. Future works may also test the reliability of laboratory-scale physical laws at coarse pixel-scale using differentiable modelling by comparing these laws and hypotheses with the relationships learned by physics-informed machine learning models \cite{wang2022discovering}. Finally, new laws that govern groundwater flow within undisturbed soils at larger scales may be discovered using geophysical imaging and system identification techniques \cite{mitchell2011inversion,terpin2024learning,guan2024identifying}.

Alternative topography-based indices such as upgradient recharge \cite{lopes2006water}, regional geomorphic indices \cite{ali2014comparison}, convergence index, and river density \cite{naghibi2020application}, could be tried in future machine learning models since they can have emergent process-oriented causal relationships with WTD. For example, \citeA{forster1988groundwater} showed that the WTD pattern is highly sensitive to the concavity or convexity of hillslopes. Certain ecological markers could also be associated with WTD and constraining groundwater models \cite{sommer2011resilience}. For example, in Nevada, \citeA{lopes2006water} noted that the Greasewood shrub species cannot grow unless the water table is less than 15 meters deep. Further studies on how different species uptake water and in-depth remotely sensed plant location data could revolutionize our knowledge of WTD. With the massive increase in the availability of satellite data, which can capture the groundwater-dependent features on the land surface at a fine spatial-temporal resolution, future works could further leverage these data to enhance the physical realism and overall strength of groundwater models. While the use of a suite of input indices, millions of pixels of data, and emergent constraints could be straightforward for machine learning models (as done in our paper), such information may be difficult to utilize in physically-based models.  

The approaches stated in this section could help us inch closer to reliable WTD estimations, but the bottom line is that we need more reliable and trustworthy observations of WTD which are precise and truly representative of the water table at a given place. Alternatively, some additional information which could guide us on the usefulness of the real observations, such as labelling confined versus unconfined observational wells as was used in our paper (see Section \ref{sec:data_filtering}), may increase certainty. Furthermore, future works could incorporate water usage datasets to help predictions follow the natural drivers of WTD while also accounting for or removing pumping biases \cite{huang2018reconstruction,wada2014global,hoekstra2012global,khan2023global,hofste2019aqueduct,janssen2021assessment}. More importantly, the groundwater modelling community urgently requires extensive global observations of WTD at aquifers with different geology located along non-floodplains and particularly steep mountainous regions. Before such data becomes available, all water table depth predictions along steep  regions should be regarded, at best, as educated guesses.

\section{Conclusions}
\label{sec:conclusion}

In this paper, three physically constrained machine learning models were developed to simulate 500 m resolution static WTD across the United States and Canada, by sequentially adding more than 12 million proxy observations of WTD to a large set of real well observations of WTD. In the first model (V1), only real observations of WTD were used to train the XGBoost machine learning algorithm. In V2, in addition to real observations of WTD, close to 12 million proxy observations along the shorelines of surface water bodies were used to train the XGBoost algorithm. Finally, in the third model (V3), more than 700,000 DEM-based proxy observations, which can roughly reflect WTD in mountainous landscapes, were added to the previous set of real and proxy observations in order to train the XGBoost algorithm. The spatial pattern of the three machine learning-based simulations of WTD, as well as two existing physically-based simulations of WTD, were compared against each other and evaluated against unseen (i.e. not used for model training) real and proxy observations. These comparisons and evaluations were made across the entire study region and within ten major ecoregions of North America to determine where these five simulations corroborate each other's conclusions and where each simulation followed currently available observations. The novelty of our work is to incorporate multiple sources of proxy observations to predict WTD using physically constrained machine learning models. Through a systematic development of three machine learning models and by comparing their simulations with two available physically-based models, we also evaluate the opportunities and challenges with data-driven WTD modeling.

We found that V1, V2 and V3, stronger than physically-based simulations, replicate unseen real observations of WTD, while V2 and V3 reasonably simulates unseen locations of permanently wet surface water bodies, in most of  North America's ecoregions. All five machine-learning and physically-based simulations performed poorly in predicting WTD along the most arid and/or high-elevation ecoregions. Pixel-by-pixel comparisons of all five simulations showed that the two physically-based models provided vastly different simulations of WTD when compared to each other and V1/V2. V3, on the other hand, closely followed the overall patterns of the two physically-based simulations while reasonably matching proxy and real observations of WTD. Opening the black box of our three machine learning models showed that V3 strongly followed classical groundwater-surface water interaction models due to topographic index strongly controlling the spatial pattern of V3 predictions, particularly at the local scale.      

Regardless of our extensive evaluation against unseen real and proxy observations, and our fairly promising results, as compared to available literature, we remain skeptical of available large-scale predictions of WTD, particularity along steep mountainous regions. Crucially, the verifiability of new, potentially promising models remains limited by our incomplete grasp of groundwater processes at large pixel scales and the pervasive influence of deeply ingrained observational biases. Yet, we remain optimistic given several future directions, including developing novel ecological or emergent indices, incorporating more carefully taken/processed observations, gaining new insights on the nonlinear functional relationships among climatological, topographical, and geological variables in shaping static WTD, and expanding the ways in which such insights can be incorporated into new models.

\acknowledgments
The interactive pixel-by-pixel maps of the three machine learning simulations of WTD, developed in this study, can be found at: \url{https://hgs4wm.eoas.ubc.ca/products}. We thank Scott Jasechko for his constructive comments. The WTD simulation data from \citeA{de2015high} is publicly available at \url{https://datacommons.cyverse.org/browse/iplant/home/shared/commons_repo/curated/DeGraaf_data_comparison_lateral_groundwater_flows_2022}. We thank Ying Fan for providing us with the WTD simulations from \citeA{fan2013global}. Joseph Janssen contributed to conceptualization, data curation, formal analysis, investigation, methodology, software, validation, visualization, writing - original draft, and writing- review \& editing. Ardalan Tootchi contributed to conceptualization, data curation, formal analysis, investigation, validation, visualization, and writing- review \& editing. Ali Ameli contributed to conceptualization, writing- review \& editing, supervision, and funding acquisition. This work was supported by the Environmental and Climate Change Canada grant awarded to Ali Ameli and the Canadian Statistical Sciences Institute (CANSSI) grant awarded to Ali Ameli. Joseph Janssen also received the Government of Canada's NSERC PhD Scholarship. All data is available upon request and the code is available at \url{https://github.com/HydroML/WTD}.

\appendix

\section{Canadian well reports}

\subsection{Quebec}
Data was downloaded from the \href{https://www.donneesquebec.ca/recherche/dataset/eau-souterraines-sih-index/resource/796a5b82-b1a6-41e2-a687-e37a171d8c6d}{\textit{Données Quebec}} database. This database is titled as the hydrogeological database. It includes all documented well drillings across the province.
It included no attributed or information classifying the producing aquifer type (confined/unconfined). Similar to many other datasets, Quebec's hydrogeological dataset represents the static level and does not include any indication on current water use of wells.

\subsection{Ontario}
Groundwater and water table depth data was downloaded from \href{https://data.ontario.ca/dataset/well-records}{WWIS2 database}. The Static Water Level attribute in WWIS2 is chosen to signify the depth of the water table. The database encompasses various features of the well of record, including casing depths, water found depth, static level, and some details regarding pumping tests.

\subsection{Nova Scotia}
Nova Scotia's Well Logs were acquired from the Government of Nova Scotia Government website. Most of the info was accessed through the \href{https://novascotia.ca/nse/groundwater/}{\textit{Department of Environment and Climate Change}} and the \href{https://novascotia.ca/natr/meb}{\textit{Department of Natural Resources and Renewables}} web pages. These databases include Well depths, casing, bedrock depth, statics water level, yield, water use and information on location and driller company. To focus on unconfined aquifer well readings, wells with static water level deeper than bedrock were excluded.

\subsection{Alberta}
Groundwater database for the Alberta Province was acquired from the Government of Alberta web page and \href{http://groundwater.alberta.ca/WaterWells/}{\textit{Interactive GIS Platform}}. A large number of  wells with static water level of zero were masked out. These consist almost 170,000 of wells in the Alberta dataset. Most of them have no information on drilled depth or perforation depth. Among these wells (WTD=0) Only those for which the drilled depth and perforation depths were also documented as zero. It consists around 30,000 wells out of 280,000 wells in the provincial database.

\subsection{British Columbia}
Water table depth across the British Columbian was extracted from the \href{https://apps.nrs.gov.bc.ca/gwells}{\textit{BC Groundwater Database}}. The database is in Excel format and it includes information on water use, total drilled depths, casing depths and diameter, bedrock depths, static water level, artesian flows (where available), and many more.
Static water levels were filtered for depths below 100 meter, and static water levels deeper than bedrock depths. Also, wells with artesian flow were also excluded.

\subsection{Saskatchewan}
Groundwater database was acquired through direct correspondences with Water Security Agency. Depths were converted from ft to m. Similar to Nova Scotia, flowing wells as well as those with static water levels shallower than screening depths were assumed to be located in confined aquifers and therefore masked out .

\subsection{New Brunswick}
The New Brunswick \href{https://www.elgegl.gnb.ca/0375-0001/index.aspx?goto=https%3a%2f%2fwww.elgegl.gnb.ca%2f0375-0001%2fpointRadiusSearch.aspx%3fqueryType%3d3%26userType%3d3%26provinceWide%3d1}{\textit{Groundwater Level Dataset}} was accessed to extract WTD observations. This dataset was  in Excel format and includes different attributes such as water use, pumping test data, drilled depth as well as the Initial Water Depth (IWD), bedrock depth, and several more. The IWD was assumed to represent the static water level. Wells with IWDs deeper than bedrock depth were excluded as they were assumed to be in confined aquifers.. 

\subsection{NewFoundland and Labrador}
Based on correspondence with Government of The Newfoundland and Labrador (NL), the NL does not keep and update a database of the groundwater levels. Data on static water levels across the Province were downloaded from the \href{https://gin.gw-info.net/service/api_ngwds:gin2/en/downloadmanager/dataset.html?package=waterwells}{\textit{GIN database}} 

\subsection{Manitoba}
Manitoba's Department of Environment was contacted to inquire about documented Provincial datasets. The groundwater dataset, however, was not accessible for public use. As a result, all available \href{https://gin.gw-info.net/service/api_ngwds:gin2/en/downloadmanager/dataset.html?package=waterwells}{\textit{GIN database}} in Manitoba was downloaded and used for the study. 

\subsection{Yukon}
Similar to a few other provinces and territories, groundwater level data across the Yukon territory was accessed and downloaded from the \href{https://gin.gw-info.net/service/api_ngwds:gin2/en/downloadmanager/dataset.html?package=waterwells}{\textit{GIN database}}.

\bibliography{bibliography}

\end{document}